\documentclass[sigconf]{aamas} 

\usepackage{balance} 

\usepackage{graphicx}

\usepackage{amsmath}
\usepackage{amsfonts}
\usepackage{xcolor}

\usepackage{amsthm}
\usepackage{mathtools}

\setcopyright{ifaamas}
\acmConference[AAMAS '23]{Proc.\@ of the 22nd International Conference
on Autonomous Agents and Multiagent Systems (AAMAS 2023)}{May 29 -- June 2, 2023}
{London, United Kingdom}{A.~Ricci, W.~Yeoh, N.~Agmon, B.~An (eds.)}
\copyrightyear{2023}
\acmYear{2023}
\acmDOI{}
\acmPrice{}
\acmISBN{}

\acmSubmissionID{275}

\title[The Importance of Credo in Multiagent Learning]{The Importance of Credo in Multiagent Learning}

\author{David Radke}
\email{dtradke@uwaterloo.ca}
\affiliation{%
  \institution{University of Waterloo}
	\city{Waterloo, ON}
	\country{Canada}
}

\author{Kate Larson}
\email{kate.larson@uwaterloo.ca}
\affiliation{%
  \institution{University of Waterloo}
	\city{Waterloo, ON}
	\country{Canada}
}

\author{Tim Brecht}
\email{brecht@uwaterloo.ca}
\affiliation{%
  \institution{University of Waterloo}
	\city{Waterloo, ON}
	\country{Canada}
}


\newcommand{\BibTeX}{\rm B\kern-.05em{\sc i\kern-.025em b}\kern-.08em\TeX}

\begin{document}


\pagestyle{fancy}
\fancyhead{}



\begin{abstract}
We propose a model for multi-objective optimization, a \emph{credo}, for agents in a system that are configured into multiple groups (i.e., teams).
Our model of credo regulates how agents optimize their behavior for the groups they belong to.
We evaluate credo in the context of challenging social dilemmas with reinforcement learning agents.
Our results indicate that the interests of teammates, or the entire system, are not required to be fully aligned for achieving globally beneficial outcomes.
We identify two scenarios without full common interest that achieve high equality and significantly higher mean population rewards compared to when the interests of all agents are aligned.
\end{abstract}


\maketitle

\newcommand{\inteam}{\nu}
\newcommand{\outteam}{\overline{\nu}}

\newcommand{\selfw}{\psi}
\newcommand{\teamw}{\phi}
\newcommand{\sysw}{\omega}



\newcommand*{\ourmean}[1]{\overline{#1}}

\newcommand{\oursubsub}[1]{\subsubsection{#1}}

\newcommand{\optional}[1]{\textcolor{YellowGreen}{#1}}

\newcommand{\todo}[1]{\textcolor{red}{\textbf{#1}}}

\newcommand{\done}[1]{\textcolor{orange}{\st{#1}}}

\newcommand{\timtext}[1]{\textcolor{black}{#1}} 

\newcommand{\oldtext}[1]{\textcolor{cyan}{#1}}

\newcommand{\heading}[1]{\vspace{3pt}\noindent\textbf{#1 }}

\newcommand{\mysubsection}[1]{\vspace{3pt}\noindent\textbf{#1 }}

\newcommand{\newtext}[1]{\textcolor{blue}{#1}}

\newcommand{\vitem}{\vspace{-5pt}\item}

\newenvironment{guideline}{\vspace{0pt} \noindent \hrulefill \\ \emph{\bf \textcolor{blue}{GUIDELINE:}} \it }{\\ \vspace{-5pt} \hrule}

\newcommand{\squishbegin}{
 \begin{list}{$\bullet$}
  { \setlength{\itemsep}{0pt}
     \setlength{\parsep}{1pt}
     \setlength{\topsep}{1pt}
     \setlength{\partopsep}{0pt}
     \setlength{\leftmargin}{1.5em}
     \setlength{\labelwidth}{1em}
     \setlength{\labelsep}{0.5em} 
  } 
}

\newcommand{\squishtwobegin}{
 \begin{list}{$-$}
  { \setlength{\itemsep}{1pt}
     \setlength{\parsep}{1pt}
     \setlength{\topsep}{1pt}
     \setlength{\partopsep}{0pt}
     \setlength{\leftmargin}{1.5em}
     \setlength{\labelwidth}{1em}
     \setlength{\labelsep}{0.5em} 
  } 
}

\newcommand{\squishend}{
  \end{list}  
}

\newcommand{\experiment}{\vspace*{4pt}\noindent\textbf{Experiment Setup:\hspace{0.4em}}}
\newcommand{\experimentend}{}

\newcommand{\moveup}{\vspace{-8pt}}
\newcommand{\movecaptionup}{\vspace{-20pt}}
\newcommand{\movecaptionuptab}{\vspace{-17pt}}
\newcommand{\colfigwidth}{0.90\columnwidth}


\newcommand{\btable}[1]{\begin{table}[#1] \begin{center} }
\newcommand{\etable}[2]{\end{center} \vspace{-5pt} \caption{#2} \label{#1} \vspace{-15pt}\end{table}}

\newcommand{\wbtable}[1]{\begin{table*}[#1] \begin{center} }
\newcommand{\wetable}[2]{\end{center} \caption{#2} \label{#1} \end{table*}}

\newcommand{\xfigure}[5]{\begin{figure}[#1] \begin{center} \leavevmode \epsfxsize=#2 \epsfbox{#3} \end{center} \vspace{-12pt} \caption{#5} \label{#4} \end{figure}}

\newcommand{\xfigurewide}[5]{\begin{figure*}[#1] \moveup \begin{center} \leavevmode \epsfxsize=#2 \epsfbox{#3} \end{center} \movecaptionup \caption{#5} \label{#4} \end{figure*}}

\newcommand{\yfigure}[5]{\begin{figure}[#1] \begin{center} \leavevmode \epsfysize=#2 \epsfbox{#3} \end{center} \caption{#5} \label{#4} \end{figure}}

\newcommand{\xyfigure}[6]{\begin{figure}[#1] \begin{center} \leavevmode \epsfxsize=#2 \epsfysize=#3 \epsfbox{#4} \end{center} \caption{#6} \label{#5} \end{figure}}

\newcommand{\bfigure}[1]{\begin{figure}[#1]}
\newcommand{\efigure}[2]{\vspace{-8pt} \caption{#2} \label{#1} \end{figure}}

\section{Introduction}
\label{sec:intro}



    
    
    

    
    

Humans have evolved the inherent ability to cooperate and organize into teams.
Some hypothesize that this has significantly supported our path of achieving higher intelligence~\cite{Rand2013HumanC,Tomasello2012TwoKS}.
Organizational Psychology and Biology have studied the structures and behavioral conditions under which people tend to be most efficient~\cite{Sundstrom1990WorkTA}.
People tend to organize themselves into ``teams-of-teams" within a larger system that are not in zero-sum competition, improving self identification and clarity of goals within a smaller group~\cite{Mathieu2001MTSs}.
Today, these teams are present at various levels of complexity in order to survive, compete in sports, or complete tasks.

Wayne Gretzky, a former ice hockey player known as \emph{The Great One}, describes a successful team as requiring ``each and every [player] helping each other and pulling in the same direction".
This statement, however, raises a number of questions.
Do players on successful teams only optimize for the goals of their team?
Is this strategy the best way for teams to achieve success?
If not, under which conditions does optimizing for an alternative goal help or hinder team success, and can incentives for individual goals actually promote behavior that is beneficial to the team?
In this work, we analyze how the performance of teams are impacted when agents may have different preferences they optimize their behavior for.

Multiagent Reinforcement Learning (MARL) has achieved impressive results in competitive two-team zero-sum settings such as capture the flag~\cite{Jaderberg2019HumanlevelPI}, hide-and-seek~\cite{Baker2020EmergentTU}, and Robot Soccer~\cite{Kitano1997RoboCupTR}, where teammates optimize for the goals of their team.
The recent call to make cooperation central to the development of AI places emphasis on understanding the mechanisms behind teamwork beyond just  competition~\cite{DafoeNature2021,Dafoe2020OpenPI} and to adapt findings from Organizational Psychology~\cite{Andrejczyk2017Concise}.
In MARL, agents learning to cooperate often build common interest by sharing exogenous rewards~\cite{Anastassacos2021CooperationAR,Baker2020EmergentRA}; however, purely pro-social agents may not be possible when considering agents designed by different manufacturers or hybrid AI/human populations.
Agents in these settings may have some self-interest for personal goals;
therefore, it is important to understand how and when cooperation can be supported in systems where agents may optimize for multiple objectives.

We introduce agent \emph{credo}, a model which regulates how agents optimize for multiple objectives in the presence of teams.
The noun credo, defined as ``the beliefs or aims which guide someone's actions"~\cite{OxfordDictionary}, fits appropriately to describe our model of how agents aim to optimize for goals.
We analyze credo in mixed-motive social dilemmas popular in recent MARL research on cooperation~\cite{Leibo2017MultiagentRL,SSDOpenSource}.
We discover multiple situations in which, despite some selfish preferences, certain credo configurations significantly outperform populations where the interests of all agents are aligned.
This work makes the following contributions:

\begin{itemize}
    \item We formally define a model of credo for multiagent teams.
    
    \item We study how the incentive structures of social dilemmas depend on agents' credo and environmental variables. 
    
    \item With learning agents, we empirically evaluate the impact of credo and find that a population can achieve over 30\% higher rewards if agents partially optimize for personal or fragmentary goals instead of goals shared by the entire population.
    
    
    
    
    
\end{itemize}



\section{Related Work}
\label{sec:related_work}




    

\noindent{\bf Human and Animal Teamwork:} 
The notion of separate entities working together to form a \emph{team} is observed across various organisms and levels of complexity.
Animals such as ants, hyenas, and whales have also been shown to display similar teamwork characteristics which increase their outputs beyond any individual~\cite{Anderson2003TeamworkIA,Anderson2001TeamsIA}.
Humans have been shown to evolve with an inherent bias towards teamwork and cooperation~\cite{Tomasello2013OriginsOH,Herrmann2007HumansHE}.
This ability to cooperate is hypothesized to have significantly contributed to our development of our intelligence and ability to learn complex skills~\cite{Tomasello2012TwoKS,Richerson1998EvolutionOfHuman}.
Humans have developed systems of ``teams-of-teams" often observed in industry, military, and sports teams that must be understood by examining between-team behavior~\cite{DeChurch2010PerspectivesTW}.

Multiteam Systems (MTSs)~\cite{Mathieu2001MTSs,Zaccaro2012MultiteamSA}, a sub-field of Organizational Psychology, specializes in understanding characteristics of these systems which lead to a larger well-functioning organism~\cite{Luciano2018MultiteamSA,Davison2012CoordinatedAI}.
MTSs research relies on human user studies which often leads to conflicting results regarding the importance of component team boundary status, connections between teams, goal type, and within-team alignment~\cite{Zaccaro2020MultiteamSA}.
A common theme across many research areas in MTSs is the idea of social identification, or how agents perceive their goals~\cite{Porck2019SocialII,Sundstrom1990WorkTA}.
Team members may need to balance tendencies for their own personal goals with the goals of their team or the entire system~\cite{Wijnmaalen2019IntergroupBI,Carter2019BestPF}.
Our model of credo formalizes how RL agents manage tension between different goals, partly inspired by the population structure and partially shared objectives of MTSs.


\noindent{\bf Teamwork in AI:} In AI, the concept of multiple non-conflicting teams that compose a larger system has been primarily explored for task completion~\cite{Grosz1996CollaborativePF,Tambe1997TowardsFT}.
This work built on years of research studying ideas of within-team coordination, such as how agents share plans~\cite{Grosz1988PlansFD}, beliefs~\cite{Pollack1986AMO,Pollack90plansas}, and intentions~\cite{Grosz1996CollaborativePF}.
Tambe~\cite{Tambe1997TowardsFT} defines a model for a system of teams that can work together towards a common goal; however, that work lacked the ability to generalize.
These early models assume teammates have full common interest.

Recent work on teams in AI has focused on ad hoc teamwork~\cite{Stone2010AdHA,Macke2021ExpectedVO,Mirsky2020APF,Durugkar2020BalancingIP,Wang2019EvolvingIM}, 
teams in competition~\cite{Ryu2021CooperativeAC,Ryu2020MultiAgentAW}
or coordination problems~\cite{Jaques2019SocialIA}.
For teams in competition, building common interest among teammates often translates to fully sharing the rewards of the team~\cite{Baker2020EmergentTU,Berner2019Dota2W}.
In social dilemmas, reward sharing is often used to align the interests of two~\cite{Anastassacos2021CooperationAR}, a subset~\cite{Radke2022Exploring,Baker2020EmergentRA}, or an entire population of agents~\cite{McKee2020SocialDA}.
Reward \emph{gifting} allows agents to incentivize others, though manifests extra reward and violates budget-balance~\cite{hostallero2020inducing,yang2020learning}.
These approaches have become standard in mixed-motive AI research to provide agents both an extrinsic reward from their environment and intrinsic reward from their peers~\cite{prosocialStag2018,Hughes2018InequityAI,Jaques2019SocialIA,Wang2019EvolvingIM}.
However, recent work for single group task completion has found their best results when agents do not only optimize for their group's preference, but also consider their individual preferences~\cite{Durugkar2020BalancingIP}.

We expand the study of multiagent team structures and multi-goal optimization to define credo, a model of how agents optimize for multiple objectives.
The model of teams we implement in our analysis is inspired by early work in both AI and MTSs.
We define credo to regulate the amount of utility common interest among teammates and between teams.
We base our analysis around similar concepts to MTSs using learning agents, such as the impact of interaction types, where cooperation happens, and under which conditions agents learn to cooperate.




\section{Model of Credo with Teams}
\label{sec:teams_model}

We model our base environment as a stochastic game $\mathcal{G}=\langle N, S, \\ \{A\}_{i\in N}, \{R\}_{i\in N}, P, \gamma, \Sigma \rangle.$
$N$ is our set of all agents that learn online from experience and $S$ is the state space, observable by all agents, where $s_i$ is a state observed by agent $i$.
$A = A_1\times \ldots \times A_N$ is the joint action space for all agents where $A_{i}$ is the action space of agent $i$.
$R = R_1 \times \ldots \times R_N$ is the joint reward space for all agents where $R_{i}$ is the reward function of agent $i$ defined as $R_i: S \times A\times S \mapsto \mathbb{R}$, a real-numbered reward for taking an action in an initial state and resulting in the next state.
$P:S\times A\mapsto \Delta(S)$ represents the transition function which maps a state and  joint action into a next state with some probability and $\gamma$ represents the discount factor so that $0 \leq \gamma < 1$.
$\Sigma$ represents the policy space of all agents, and the policy of agent $i$ is represented as $\pi_i:S \mapsto A_{i}$ which specifies an action that the agent should take in an observed state.\footnote{We can also allow for randomized policies.}

We introduce agent credo, a model to regulate how much an agent optimizes for different reward components it has access to. 
In this paper, we use ``common interest" in the utility sense, referring to when agents share exogenous rewards instead of sharing views.
A \emph{team} is a subset of agents which have some degree of common interest for team-level goals.
Given a population, multiple teams with different preferences and interests may co-exist that are not in zero-sum competition.
The collection of all teams is referred to as a team \emph{structure}.
We denote the set of all teams as $\mathcal{T}$, the teams agent $i$ belongs to as $\mathcal{T}_i$, and a specific team as $T_{i} \in \mathcal{T}_i$.

Our goal is to relax the modelling assumption that teammates are bound through full common interest~\cite{Radke2022Exploring,Jaderberg2019HumanlevelPI,Baker2020EmergentTU,chung2021map} to study how different credos impact a system of learning agents.
For example, an agent may optimize their policy for the performance of one or multiple teams, while also being somewhat oriented towards its own personal goals.
We represent these guiding principles by decomposing the reward any agent $i$ may receive from the environment into three components: their individual exogenous reward $IR_{i} = R_i$, the rewards $i$ receives from each team they belong to $TR_{i}^{T_i} \forall T_i \in \mathcal{T}_i$, and the reward $i$ receives from the system of $N$ agents $SR_i$.
$TR_{i}^{T_i}$ and $SR_i$ can be implemented with any function to aggregate and distribute rewards amongst multiple agents.

We define credo to be a vector of \emph{parameters}, $\mathbf{cr}_i$, where the sum of all parameters is 1.
The credo of an agent is represented $\mathbf{cr}_i = \langle \selfw_{i}, \teamw_{i}^{T_1}, \dots, \teamw_{i}^{T_{|\mathcal{T}|}}, \sysw_{i} \rangle$, where $\selfw$ is the credo parameter for $i$'s individual reward $IR_i$, $\teamw_{i}^{T_i}$ is the credo parameter for the reward $TR_{i}^{T_i}$ from team $T_i \in \mathcal{T}_i$, and $\sysw_i$ is the credo parameter for the reward $i$ receives from the system $SR_i$.
The parameter notation is organized by increasing order of group size, so that  $\mathbf{cr}_i = \langle$self$, \dots,$ teams$, \dots,$ system$ \rangle$, where |self| < |teams| $\leq$ |system|.
Agent $i$'s credo-based reward function $R^{\mathbf{cr}}_i$ is calculated as:

\begin{equation}
    R^{\mathbf{cr}}_i = \selfw_i IR_i + \sum_{T_i \in \mathcal{T}_{i}} \teamw_{i}^{T_i} TR_{i}^{T_i} + \sysw_i SR_i ,
    \label{eq:credo_calc}
\end{equation}



The environment in our analysis consists of a stochastic game with a model of team structure
 $\langle \mathcal{G}, \mathcal{T} \rangle$.
A particular case is where agents belong to exactly one team, which we study in the rest of this paper.
Formally, $\mathcal{T}$ is a partition of the population into disjoint teams,  $\mathcal{T} = \{ T_i | T_i \subseteq N, \cup T = N, T_i \cap T_j = \emptyset \forall i, j \}$.
This team structure simplifies the credo vector for each agent to be $\mathbf{cr}_i = \langle \selfw_{i}, \teamw_{i}, \sysw_{i} \rangle$ where $\teamw_i$ is the credo parameter for $i$'s team.


Any function can be used to calculate $IR_i$, $TR_i^{T_i}$, or $SR_i$ in our model.
We implement functions to be consistent with building common interest between learning agents, similar to past work~\cite{Wang2019EvolvingIM,Baker2020EmergentTU,Jaderberg2019HumanlevelPI}.
$IR_i$ is defined to be the agent's normal exogenous reward $R_i$.
Their team reward is defined as $TR_i^{T_i}:S\times A_i \times S \mapsto \mathbb{R}$, so that:

\[ TR_i^{T_i}=\frac{\sum_{j\in T_i} R_j(S, A_j, S)}{|T_i|},\]

\noindent
where teammates share their rewards equally.
The system-wide reward is defined as $SR_i:S\times A_i \times S \mapsto \mathbb{R}$ so that:

\[ SR_i=\frac{\sum_{j\in N} R_j(S, A_j, S)}{|N|},\]

\noindent
the mean reward of all $N$ agents.
The final credo-based reward for agent $i$, $R^{\mathbf{cr}}_{i}$, is calculated using Equation~\ref{eq:credo_calc} and these functions.

As is standard in many MARL problems, agents are trained to independently maximize their rewards. 
In particular, at time $t$ each agent $i$ selects some action $a_i$ which together form a joint action $a^t$.  
This action results in a transition from state $s^t$ to state $s^{t+1}$, according to the transition function $P$, and provides each agent $i$ with reward $R_{i,t}(s^t,a^t,s^{t+1})$. 
Agents seek to maximize their sum of discounted future rewards, $V_i=\sum_{t=0}^\infty \gamma^t R_{i,t}$.
Our model replaces $R_{i}$ with $R^{\mathbf{cr}}_{i}$, reconfiguring the learning problem so agents must learn behavior that maximizes their sum of discounted future credo-based rewards according to the team structure and environment.

\section{Social Dilemmas}
\label{sec:social_dilemmas}


Social dilemmas present tension between short-term individual incentives and the long-term collective interest, where agents must cooperate for higher rewards.
All agents prefer the benefits of a cooperative outcome; however, the short-term benefits of self-interested behavior outweigh those of cooperative behavior, making social dilemmas an excellent testbed for credo.
Social dilemmas and mechanisms for agents or people to overcome them has been widely studied across game theory~\cite{Santos2018SocialNC,Challet1997EmergenceOC}, economics~\cite{Ostrom1990GoverningTC,Brady1993GoverningTC}, psychology~\cite{dawes2000,Fehr2018NormativeFO}, and more recently, AI~\cite{Leibo2017MultiagentRL,Danassis2021ImprovedCB,Anastassacos2021CooperationAR}.
For our analysis, we consider intertemporal social dilemmas with active provision, defined as when cooperation carries an explicit cost~\cite{Hughes2018InequityAI}.
We implement our model of credo and teams in the Iterated Prisoner's Dilemma (IPD)~\cite{Rapoport1974PrisonersD} and Cleanup Markov game~\cite{SSDOpenSource}.

\subsection{Iterated Prisoner's Dilemma (IPD)}

In the one-shot Prisoner's Dilemma, two agents interact and each must decide to either cooperate ($C$) with or defect ($D$) on each other.
We assume there is a cost ($c$) and a benefit ($b$) to cooperating where $b>c>0$ with payoffs shown in Table 1 in the appendix.\footnote{\url{https://cs.uwaterloo.ca/~dtradke/pdfs/credo_apdx.pdf}}
If an agent cooperates, it incurs the cost $c$.
If both agents cooperate, they both benefit, each receiving a reward of $b-c$.
If one agent cooperates but the other defects, then we assume that the cooperating agent incurs the cost $c$, but the defecting agent reaps the benefit $b$ (e.g., by stealing the contribution of the cooperator).
If neither cooperate, neither benefit nor incur a cost, leading to a reward of zero for both.
The unique Nash Equilibrium is obtained when both agents defect,
since if one agent cooperates, the other agent is strictly better off defecting and receiving $b$, instead of $b-c$.

The degree to which agents optimize for different groups may change how these interactions are viewed.
The traditional assumption is that each agent acts individually and is solely focused on their own individual reward.
However, if agents have certain credos, they may optimize for entire groups they belong to instead of themselves which may have a considerable impact on how agents learn.
We can view the standard Prisoners' Dilemma in Table 1 as an instance of the game where the credo of both agents is fully self-focused ($\mathbf{cr}_i=\langle 1,0,0\rangle$).
If both agents were entirely system-focused ($\mathbf{cr}_i=\langle 0,0,1\rangle$), or on the same team and team-focused ($\mathbf{cr}_i=\langle 0,1,0\rangle$), agents would wish to take actions which maximize their shared utility, resulting in mutual cooperation.

In the IPD, this game is played repeatedly which adds a temporal component and allows agents to learn a policy over time.
Instead of just two agents, we work with a population of agents that are divided into teams \emph{a priori}.
At each timestep, agents are randomly paired with another agent, a \emph{counterpart}, that may or may not be a teammate.
Agents are informed as to what team their counterpart belongs to through a numerical signal $s_i$, though additional identity information is not shared.
Agents must decide to cooperate with or defect on this counterpart for a single round of the Prisoner's Dilemma. 
Their payoff for the interaction is their reward defined by their credo, $R^{\mathbf{cr}}_i$, based on their own and other agents' interactions.
Agents update their strategies (i.e., learn) using their direct observation $s_i$, what action they chose $a_i$, and their credo reward $R^{\mathbf{cr}}_i$.
Since only the team information of the counterpart is shared, the strategies of all agents on team $T_i$ ultimately affects how agents learn to play any member of $T_i$.
Further implementation details can be found in Appendix A.1.

\subsubsection{Equilibrium Analysis with Credo}
\label{sec:agent_credo}

We are interested in understanding the conditions under which credo may help or hinder cooperation.
Thus, as a first step we investigate the impact of credo on the \emph{stage game} of the IPD with teams.
To provide a clear comparison with the standard IPD, we take an ex-ante approach, where agents are aware of their imminent interaction and the existence of other teams, but not the actual team membership of their counterpart.
Refer to Appendix B for further details and the complete derivation of our equilibrium analysis. 

Assume a pair of agents, $i,j$, have been selected to interact at some iteration of the IPD and agent $i$ knows $j$ will be a teammate with probability $\inteam$ and a non-teammate with probability $(1-\inteam)$.
Let $\sigma_{T_i}=(\sigma_{ji},1-\sigma_{ji})$ represent $j$'s strategy profile when $j \in T_i$, where $\sigma_{ji}$ is the probability for cooperation ($C$).
Likewise, let $\sigma_{T_j}=(\sigma_{jj},1-\sigma_{jj})$ be $j$'s strategy profile when $j \in T_{j}$, any other team.

For the sake of our analysis, we make the assumption that all agents have the same credo.
We calculate the expected values of cooperation and defection in situations where agents are fully self-focused ($\mathbf{cr}_i = \langle 1, 0, 0 \rangle$), team-focused ($\mathbf{cr}_i = \langle 0, 1, 0 \rangle$), and system-focused ($\mathbf{cr}_i = \langle 0, 0, 1 \rangle$).
We then calculate the conditions in which agent $i$ has the incentive to cooperate as when the expected value of cooperation dependant on credo is greater than the expected value of defection.
After algebraic simplification, we determine agent $i$ is better off cooperating whenever:
\begin{align}
     \teamw_i \left( \inteam - \frac{2c}{b+c} \right) + \sysw_i \left( \frac{b-c}{2} \right) \geq \selfw_i c.
    \label{eq:final_requrement}
\end{align}

\noindent
Note that this is independent of the strategy profile of their counterpart, $\sigma_T$.
Whenever cooperation is the dominant strategy in a stage game, it will be supported in the repeated game.


\begin{figure}[t]
    \centering
    \includegraphics[width=0.9\linewidth]{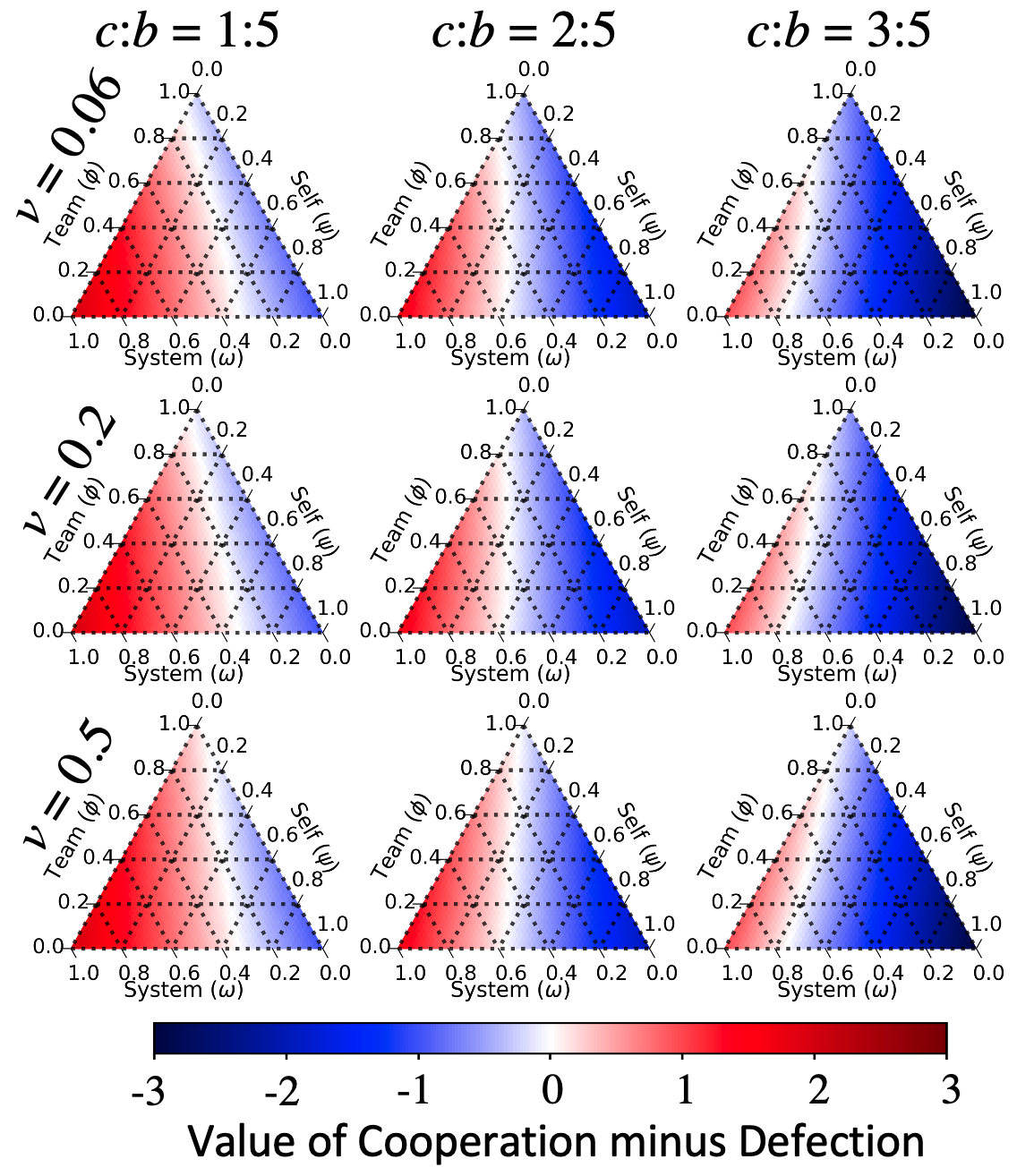}
    \caption{Impact of teammate pairing probability $\inteam$ and the cost of cooperation $c$ (benefit $b = 5$) on action incentives with credo. Red corresponds with cooperation being incentivized and blue corresponds with defection.}
    \label{fig:credo_theory}
\end{figure}


Figure \ref{fig:credo_theory} shows the expected reward value of cooperation minus the expected reward value of defection, solving Equation \ref{eq:final_requrement} for $|\mathcal{T}| = 5$ teams.
Each triangle shows the results for the linear combination of agent credo composed of self- ($\selfw$; right axis), team- ($\teamw$; left axis), and system-focused ($\sysw$; bottom axis) parameters (increments of 0.02).
The colors indicate which action has higher expected value, defection (blue) or cooperation (red), computed as the difference of their expected values (white represents equality).

Each row of plots represents different values of $\inteam$, the likelihood of being paired with a teammate.
The remaining probability $1-\inteam$ is spread across the $|\mathcal{T}|-1$ teams uniformly.
With five teams, these values of $\inteam$ represent when the chance of a counterpart being from another team is four times more likely than their own team ($\inteam = 0.06$), being from any of the five teams has equal probability ($\inteam = 0.2$), and being from the same team is four times more likely than another team ($\inteam = 0.5$).
Each column of plots represents a different cost of cooperation so that $c \in \{1, 2, 3\}$ with the benefit fixed to $b = 5$.
For our entire analysis, we increase the cost and fix the benefit since we are interested in the ratio between the cost and benefit of cooperation instead of their absolute values.

We observe less overall incentive to cooperate as the cost $c$ increases (i.e., darker blue and has more area inside the triangles).
This pattern resembles findings observed in human behavior, where the amount of cooperation depends on the size of the benefit compared to the cost~\cite{Schnell2021levels}.
Another observation is that defection is incentivized in the presence of any amount of self-focus (right axes), with the exception of one environment ($c=1$ and $\inteam = 0.5$).
Even in this scenario, defection becomes quickly incentivized as self-focus increases to $\selfw_i = 0.2$.
The experiments in Sections~\ref{sec:ipd_evaluation} show that learning agents are able to develop globally beneficial cooperative behavior in multiple settings where defection is incentivized.




\subsection{Cleanup Markov Game}



The Cleanup domain~\cite{SSDOpenSource} is a temporally and spatially extended Markov game representing a social dilemma.
Cleanup allows us to examine if our findings generalize to more complex environments since agents must learn a cooperative policy through movement and decision actions instead of choosing an explicit \emph{cooperation} action like in the IPD.
Active provision is represented in Cleanup by agents choosing actions with no associated environmental reward that are necessary before any agent achieves a reward.

Appendix A.2
contains a screenshot of the Cleanup environment with $|\mathcal{T}| = 3$ teams.
At each timestep, agents choose among nine actions: 5 movement (up, down, left, right, or stay), 2 turning (left or right), and a cleaning or punishing beam.
One half of the map contains an aquifer (or river) and the other an apple orchard.
Waste accumulates in the river with some probability at each timestep which must be cleaned by agents.
Apples spawn in the orchard with a function that accounts for the cleanliness of the river.
Agents receive a reward of +1 for consuming an apple by moving on top of them.
The dilemma exists in agents needing to spend time cleaning the river (to spawn new apples) and receiving no exogenous reward versus just staying in the orchard and enjoying the fruits of another's labor.
Agents have the incentive to only pick apples; however, if all agents attempt policy, no apples grow and no agents get any reward.
A successful society will balance the temptation to free-ride with the public obligation to clean the river.

\section{Empirical Evaluation}


In the rest of this paper, we present the setup and results of experiments in both environments using learning agents.
While our model does not require it, we assume that for all teams $T_i,T_j\in\mathcal{T}$, $|T_i|=|T_j|$ (i.e., given a team model, the teams are the same size).
This avoids complications that might arise with agent interactions if teams were of significantly different sizes and to be consistent across domains, leaving other interactions and different sized teams for future work.
We initialize $\mathbf{cr}_i$ to be the same for all agents \emph{a priori} and do not change $\mathbf{cr}_i$ within an experiment.
Since fully self-focused and system-focused credos have agents working as individuals (i.e., non-team setting) and a full group (i.e., cooperative setting), they serve as benchmarks against which we can compare the performance of other credos with teams.

\subsection{IPD Evaluation}
\label{sec:ipd_evaluation}

In the IPD, each experiment lasts $1.0 \times 10^6$ episodes.
We configure $N=25$ Deep $Q$-Learning~\cite{Mnih2015HumanlevelCT} (DQN) agents into $|\mathcal{T}| = 5$ teams of equal size.
While our team model allows for an arbitrary number of teams of any size, this work is concerned with the relationship between agent credo and environmental conditions.

An episode is defined by a set of agent interactions where each agent is paired with another agent and plays an instance of the Prisoner's Dilemma.
Agent pairings are assigned based on $\inteam$, the probability of being paired with a teammate and agents are unable to explicitly modify who they interact with, a challenging scenario for cooperation without additional infrastructure~\cite{Anastassacos2020PartnerSF}.
Each experiment is repeated five times with different random seeds.
Further implementation details are provided in Appendix A.1. 

\begin{figure}[t]
    \centering
    \includegraphics[width=\linewidth]{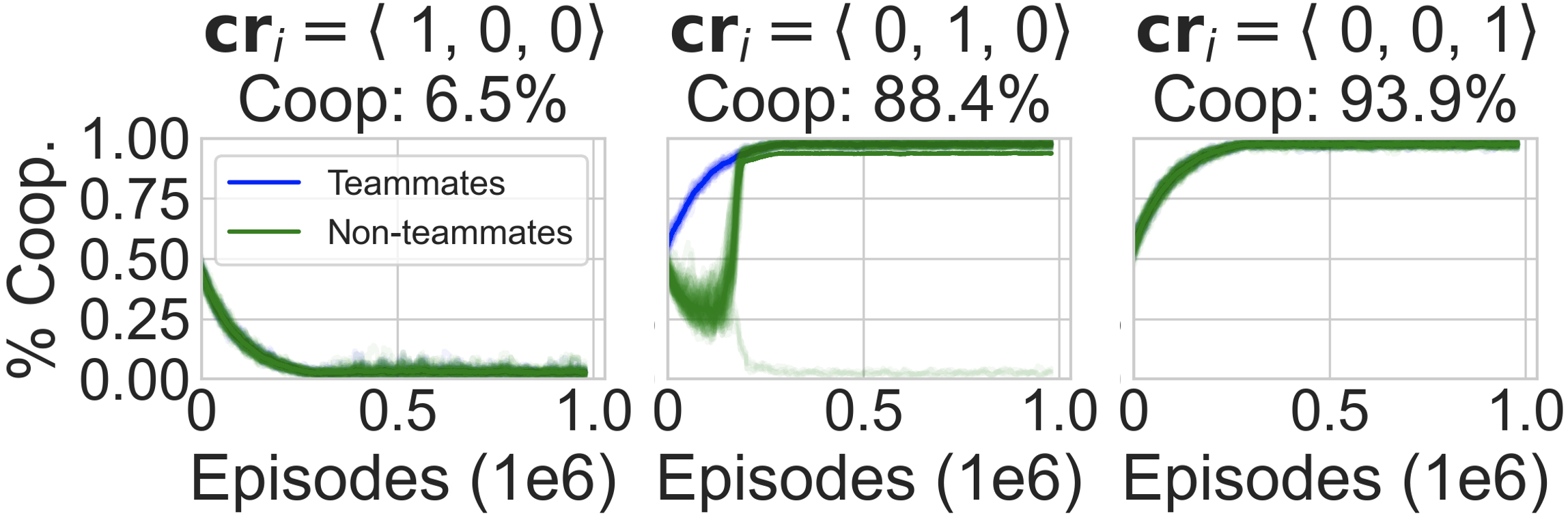}
    \caption{\textit{IPD: }Full self-, team-, and system-focused agents when $c = 1$, $b = 5$, $\inteam = 0.2$, and $|\mathcal{T}| = 5$ of five agents each.}
    \label{fig:teams_cooperate}
\end{figure}

\begin{figure*}[t]
    \centering
    \includegraphics[width=\linewidth]{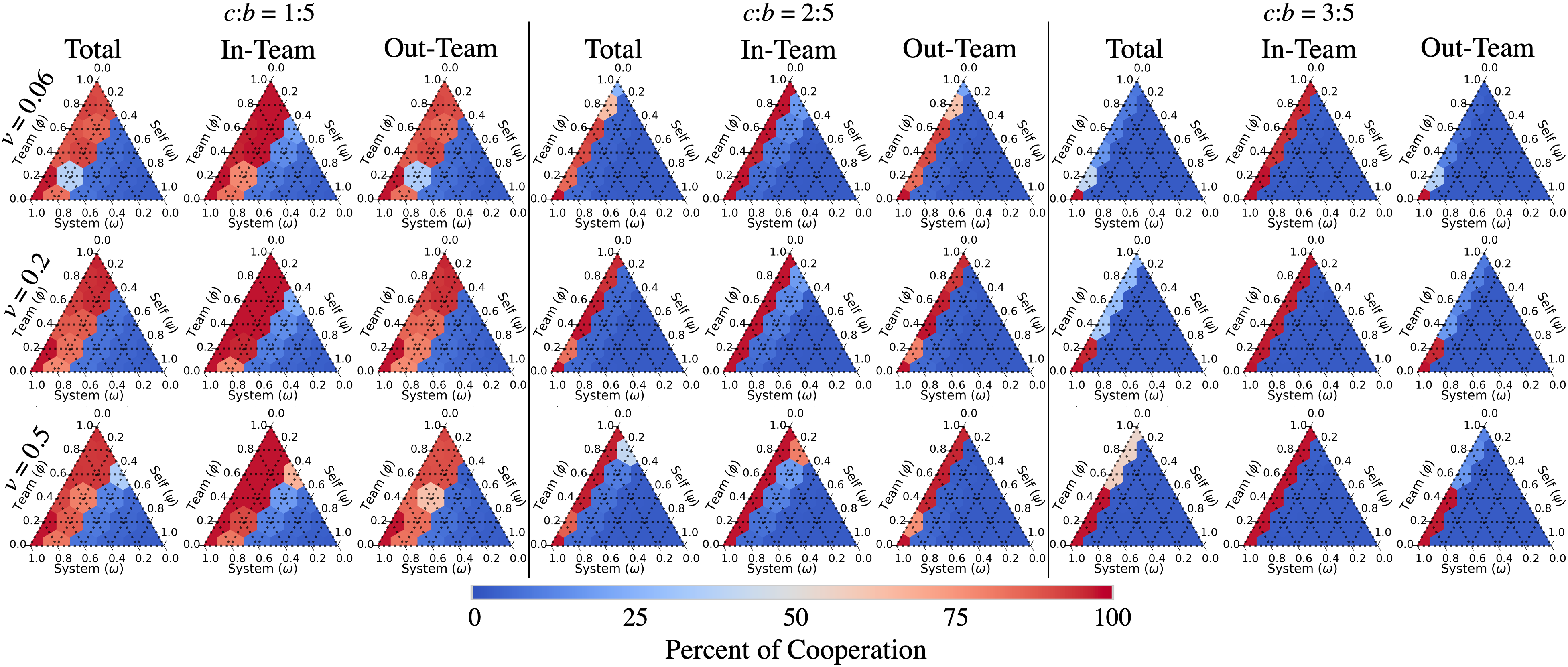}
    \caption{\textit{IPD:} Percent of total cooperation, cooperation with teammates (In-Team), and cooperation with non-teammates (Out-Team) when all agents follow the same credo. We experiment with different probabilities of being paired with a teammate $\inteam \in \{0.06, 0.2, 0.5\}$, costs of cooperation $c \in \{1, 2, 3\}$, and fix the benefit $b = 5$.}
    \label{fig:homog_credo-impact}
\end{figure*}

\subsubsection{IPD: Full-Focus Credo}
\label{sec:results}

We start by analyzing how the behavior of agents is impacted by the extreme cases of full self-focus ($\mathbf{cr}_i = \langle 1, 0, 0 \rangle$), team-focus ($\mathbf{cr}_i = \langle 0, 1, 0 \rangle$), or system-focus ($\mathbf{cr}_i = \langle 0, 0, 1 \rangle$) credo.
Figure~\ref{fig:teams_cooperate} shows our results where
the $x$-axis of each plot shows the time and the $y$-axis shows the percent of actions where agents chose to cooperate, averaged over 2,000 episode windows.
We set $c = 1$, $b = 5$, and $\inteam = 0.2$ so counterparts have equal probability of being selected from any team (since $|\mathcal{T}| = 5$).
Each line represents the behavior of one agent when paired with a teammate (blue) or a non-teammate (green).

When all agents are fully self-focused (left), $\mathbf{cr}_i = \langle 1, 0, 0 \rangle \forall i \in N$, they immediately learn defection towards all other agents (blue overlapped by green).
When agents are team-focused (middle), $\mathbf{cr}_i = \langle 0, 1, 0 \rangle$, defection is incentivized as shown in Figure \ref{fig:credo_theory}.
However, we find agents quickly identify and cooperate with their teammates and almost every agent simultaneously develops stable pro-social policies with non-teammates despite not sharing rewards.
We hypothesize this is due to a combination of reduced reward variance for actions in specific states and interactions with teammates providing a strong positive feedback signal favoring cooperation.
In the IPD, fully system-focused DQN agents will achieve the upper bound of performance since they maximize the collective rewards of all agents, equivalent to choosing cooperation.
In the right plot when agents are fully system-focused, agents do learn to cooperate with every agent regardless of team (blue overlapped by green).
While other work requires strong assumptions of behavior to steer agents towards cooperation~\cite{Anastassacos2021CooperationAR}, these results indicate that full common interest may not be required to promote cooperation across an entire population with teams.

\subsubsection{IPD: Multi-Focus Credo}

Next, we experiment with settings where agents can simultaneously partially optimize for their own, their team's, or the system's goals through their credo.
We use the same environmental settings as Figure~\ref{fig:credo_theory}, so that $\inteam \in \{0.06, 0.2, 0.5\}$, $c \in \{1, 2, 3\}$, and $b = 5$ to understand how credo and environmental parameters impact how agents learn.
We evaluate the case where all agents have the same credo parameters, that is, $\mathbf{cr}_i = \mathbf{cr}_{j} \forall i, j \in N$.



Figure \ref{fig:homog_credo-impact} shows our results for various combinations of credo with 0.2 step increments, teammate pairing probability $\inteam$, and the cost of cooperation $c$.
This creates nine different environments, each with 21 combinations of credo represented by the intersections of dotted lines from the three axes of each triangle of Figure \ref{fig:homog_credo-impact}.
In the IPD, mutual cooperation yields the highest mean population reward.
Thus, the hexagonal area around each intersection point is colored according to the global mean percent of actions which were to cooperate from blue (less cooperation) to red (more cooperation).
RL agents are able to condition their policy on the information of their counterpart's team, allowing us to observe how they learn behavior towards different groups.
For each environment, we plot the total cooperation, cooperation with teammates (In-Team), and cooperation with non-teammates (Out-Team).

Agents achieve high cooperation when they have full system-focus (left corners) and learn defection as agents become self-focused (right corners).
Despite the incentive to defect in eight of nine environments, fully team-focused agents learn cooperation in five environments, similar to the behavior in Figure~\ref{fig:teams_cooperate}.
This cooperation is robust if full team-focus can not be achieved, such as when self-focus $\selfw = 0.2$ and $c=1$.
In these settings, the rate of cooperation is higher when agents have high team-focus compared to high system-focus, though decreases as when $c=3$.
Unlike previous implementations of teams which assume agents have full common interest~\cite{Jaderberg2019HumanlevelPI,Baker2020EmergentTU,chung2021map}, our results show teammates are not required to be fully aligned to achieve good results.

This shows that teams of highly team-focused agents have the ability to support global cooperation despite incentives to defect
or some self-focus, relaxing the assumption that teammates are bound together with full common interest.
Contrary to Gretzky's belief in Section~\ref{sec:intro}, our results indicate teams still achieve good results despite some self-focus.
To understand this significance, consider situations where pure-common interest among teammates may not be guaranteed or all agents are unable to be controlled.
Shown next, results in the Cleanup domain actually improve beyond full system-focus with certain credo parameters.

\subsection{Cleanup Evaluation}

Using the Cleanup evaluation domain, we experiment with $N=6$ agents learning with Proximal Policy Optimization (PPO)~\cite{PPO2017} split into $|\mathcal{T}| = 3$ teams of two agents each.
Past work that has used the Cleanup domain typically uses $N = 5$ agents for a time of $1.6 \times 10^8$ environment steps (each episode is 1,000)~\cite{Hughes2018InequityAI,Wang2019EvolvingIM}.
We increase the population to $N = 6$ agents to allow for three equal sized teams and calculate metrics over the last 25\% of timesteps, similar to the IPD evaluation.
Agent observability is limited to a 15 $\times$ 15 RGB window centered on the agent's current location.
Teammates appear as the same color and optimize their own $R^{\mathbf{cr}}_i$ after each timestep.
Each experiment is completed eight times with different random seeds.
Further experiment details are found in Appendix A.2.


\subsubsection{Cleanup: Reward.}

\begin{figure}[t]
    \centering
    \includegraphics[width=0.8\linewidth]{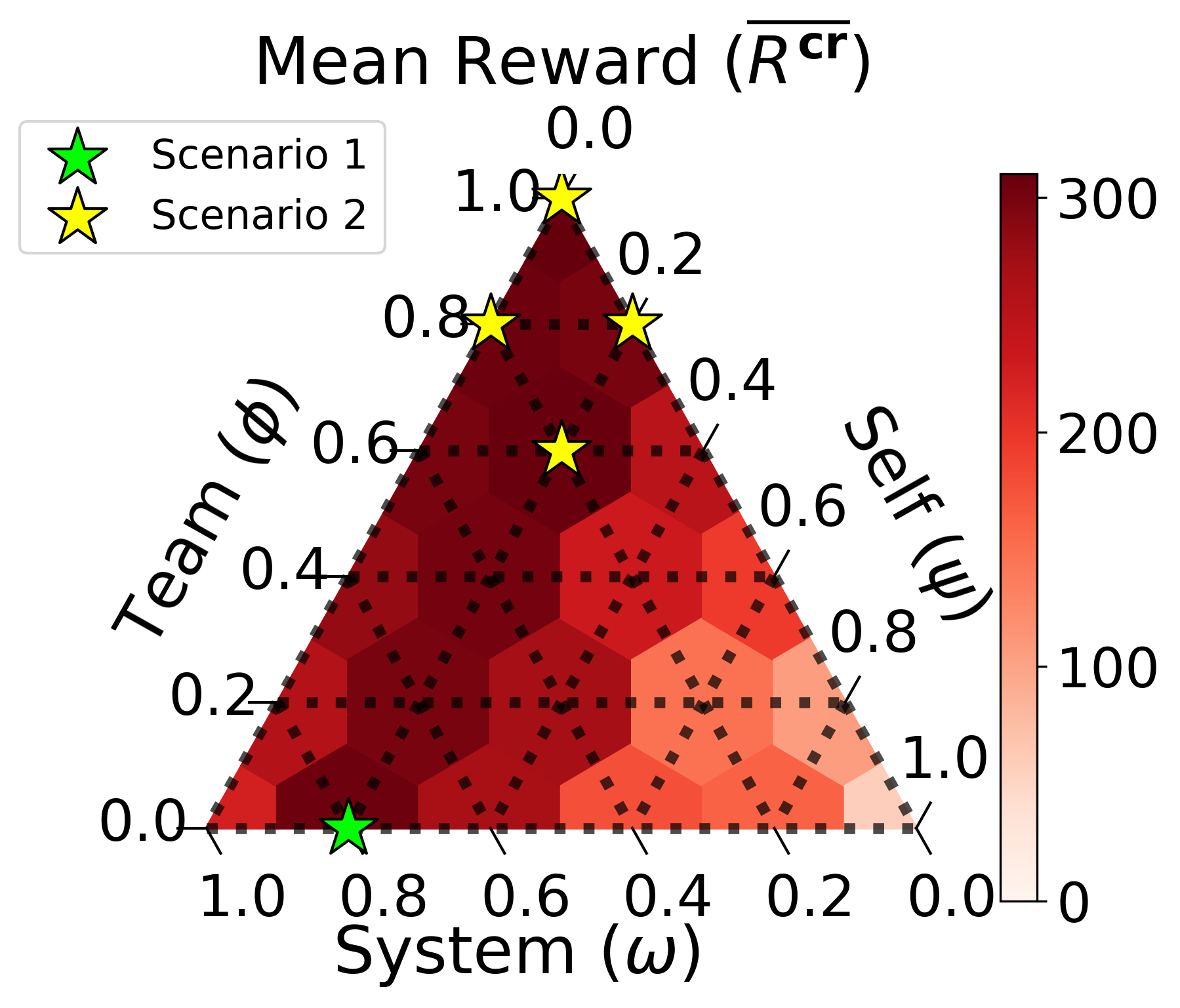}
    \caption{\textit{Cleanup:} Mean population reward for every credo in our evaluation. These experiments have $|\mathcal{T}| = 3$ teams of two agents each. The scenarios with the highest reward often have agents with slight self-focus. We identify two types of credo scenarios that achieve the highest reward, when credo has slight self-focus paired with high system-focus (green star) and when team-focus is high (yellow stars).}
    \label{fig:cleanup_results-reward}
\end{figure}

While success in the IPD relies on agents choosing to cooperate in direct interactions, success in Cleanup requires agents to coordinate and form an effective joint policy to clean the river enough for apples to grow.
Since agents must learn their own coordinated policy by taking actions in states with higher complexity, we can no longer assume the full-common interest scenario (full system-focus) will achieve the upper bound on performance.
Figure~\ref{fig:cleanup_results-reward} shows the mean credo-based population reward per-episode $\overline{R^{\mathbf{cr}}}$ for all 21 credo configurations in Cleanup, removing the subscript $i$ when referencing all agents.
The mean population reward gives insight into how well agents learn to solve the dilemma.
Each hexagon corresponds with a combination of credo, centered at the intersections of three dotted lines from each axis (self, team, and system).
Hexagons are colored according to the mean population reward from low (white) to high (red).

Fully self-focused agents fail to solve the dilemma, receiving the lowest mean population reward of any scenario.
Previous work has found that the highest rewards in Cleanup are obtained when agents optimize for reward signals from all agents (i.e., system-focus)~\cite{Wang2019EvolvingIM,McKee2020SocialDA,gemp2020d3c}.
However, we find that some self- or team-focus improves the mean reward significantly over the system-focused setting.
We divide the five highest-reward environments into two scenarios shown in Figure~\ref{fig:cleanup_results-reward}.
First, when agents with high system-focus also have slight self-focus (Figure~\ref{fig:cleanup_results-reward}, green star), and second, when agents have high team-focus relative to their other credo parameters (Figure~\ref{fig:cleanup_results-reward}, yellow stars).
These scenarios achieve at least 30\% higher mean population reward per-episode than the full common interest setting (system-focused; left corner).

\textbf{Scenario 1:}
The first scenario we examine is when highly system-focused agents have slight self-focus, $\mathbf{cr}_i  = \langle 0.2, 0.0, 0.8 \rangle$ (green star).
Agents with this credo achieve 33\% higher reward per-episode compared to a population with full common interest.
This result is comparable to a similar finding in past work~\cite{Durugkar2020BalancingIP}, where a cooperative group performs best when agents have some selfish preferences of how to complete a task.
This suggests that, despite using an entirely different domain, some amount of self-focus may consistently lead to high global performance and is worthy of more exploration.

\textbf{Scenario 2:}
With the introduction of teams-focus, we more closely examine another credo scenario that contains four of the top five experiments with high mean rewards.
The yellow stars in Figure~\ref{fig:cleanup_results-reward} show experiments when agents have high team-focus relative to their other credo parameters, specifically $\mathbf{cr}_i \in \{\langle 0.0, 1.0, 0.0 \rangle, \\ \langle 0.2, 0.8, 0.0 \rangle, \langle 0.0, 0.8, 0.2 \rangle, \langle 0.2, 0.6, 0.2 \rangle \}$.
As discussed in Scenario 1, agents with high system-focus experience a decrease in rewards when they are \emph{too} system-focused.
In Scenario 2, high team-focused agents achieve high rewards regardless if self-focus is zero or 0.2, echoing our result in the IPD that teammates are not required to have full common interest.
These insights may be useful when attempting to influence credo in settings where agents are unable to guarantee their amount of self-focus or team commitment.

\subsubsection{Cleanup: Division of Labor}

We find that agents in the highest-reward experiments often learn to divide labor and specialize to either clean the river or pick apples.
This ability to coordinate with other teams and fill roles significantly impacts the global reward.
We observe the best division of labor strategy when two agents clean the river (i.e., cleaners) and four agents pick apples (i.e., pickers).
In the following analysis, each line in Figures~\ref{fig:cleanup_policy_sys-someself} and~\ref{fig:cleanup_policy_highteam} represent the behavior of a single agent.
For example, a line labeled ``a-0/$T_0$'' represents agent \#0 belonging to team \#0.
Teammates appear as different shades of the same color ($T_0$ blue, $T_1$ red, and $T_2$ green).
In adjacent trials, agents with the same label (i.e., a-0/$T_0$) may learn different behavior, making aggregating policies from all eight trials difficult.
Therefore, we present figures from one trial representing the most commonly learned behavior in each setting.


\begin{figure}[t]
    \centering
    \includegraphics[width=0.95\linewidth]{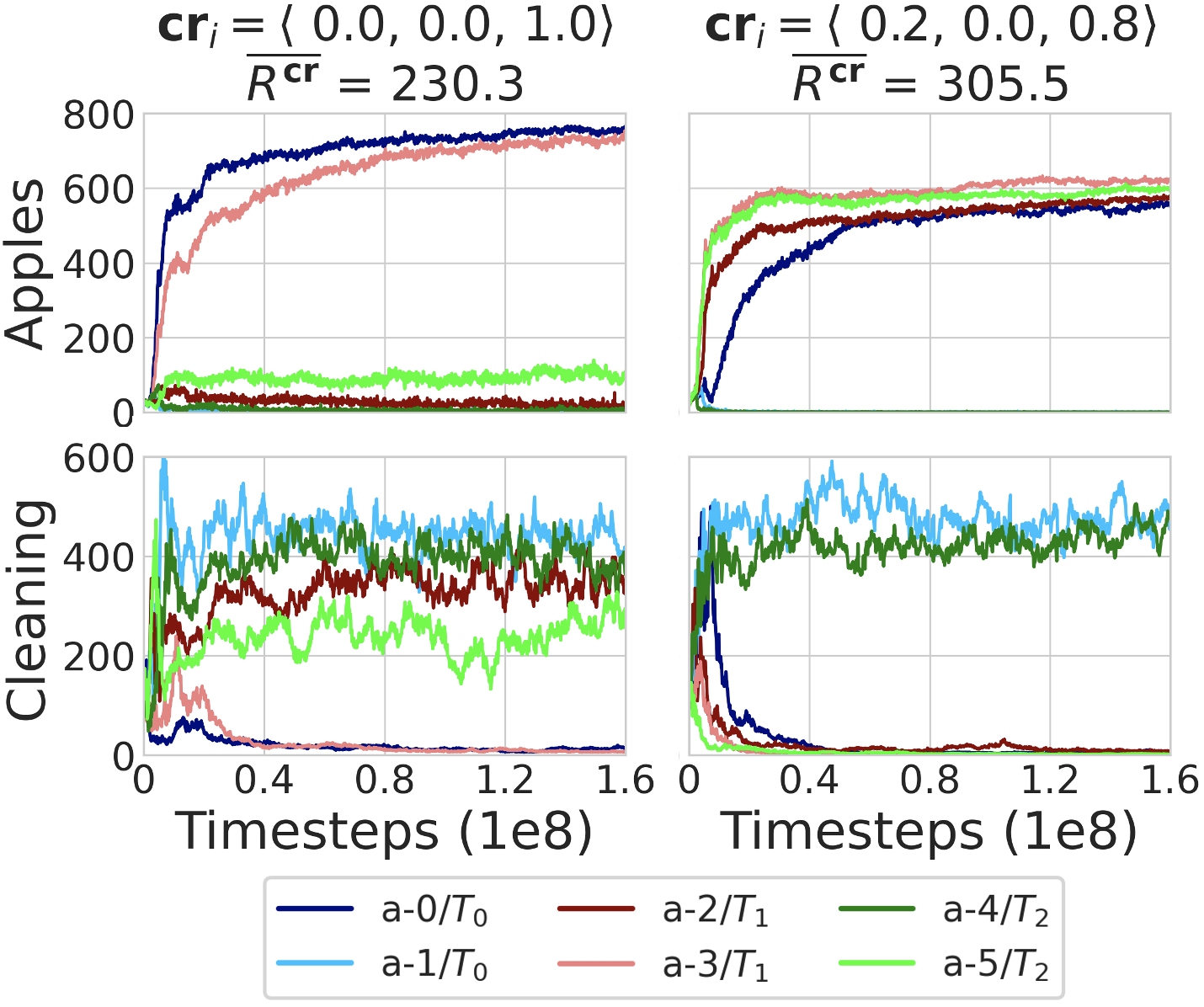}
    \caption{\textit{Cleanup:} (Scenario 1) high system-focus with slight self-focus. Agents are labeled so that ``a-0/$T_0$'' represents agent \#0 belonging to team \#0. Each column of plots shows (left) fully system-focused agents and (right; green star in Figure~\ref{fig:cleanup_results-reward}) when agents become slightly self-focused. Better division of labor strategies are learned when self-focus increases from zero to 0.2 by enticing four agents to pick apples instead of just two, leading to 33\% higher reward.}
    \label{fig:cleanup_policy_sys-someself}
\end{figure}

\textbf{Scenario 1:}
We first analyze the division of labor in Scenario 1 (green star in Figure \ref{fig:cleanup_results-reward}).
Figure~\ref{fig:cleanup_policy_sys-someself} shows the number of apples picked (top) and cleaning actions taken (bottom) by all six.
The left plots show when agents are fully system-focused ($\mathbf{cr}_i = \langle 0.0, 0.0, 1.0 \rangle$) and the right plots shows when agents become slightly self-focused ($\mathbf{cr}_i = \langle 0.2, 0.0, 0.8 \rangle$).
The full system-focused population evolves into four cleaning agents and two apple pickers, with each cleaner receiving rewards from both pickers regardless of team membership.
This amount of reward suppresses any desire for cleaning agents to learn to pick apples, causing the population to reach a local minimum.
The two apple pickers pick over 700 apples each resulting in a mean population reward of $R^{\mathbf{cr}} = 230.3$.
However, increasing the self-focus to $\teamw_i = 0.2$ (right plots) provides enough individual incentive to for four agents to pick apples and collect 600 apples each for $R^{\mathbf{cr}} = 305.5$.
Due to high system-focus, the two cleaning agents receive enough reward from all four pickers to incentivize them to continue cleaning, and the entire system achieves 33\% higher reward by escaping the previous local minimum.

\begin{figure}[t]
    \centering
    \includegraphics[width=0.95\linewidth]{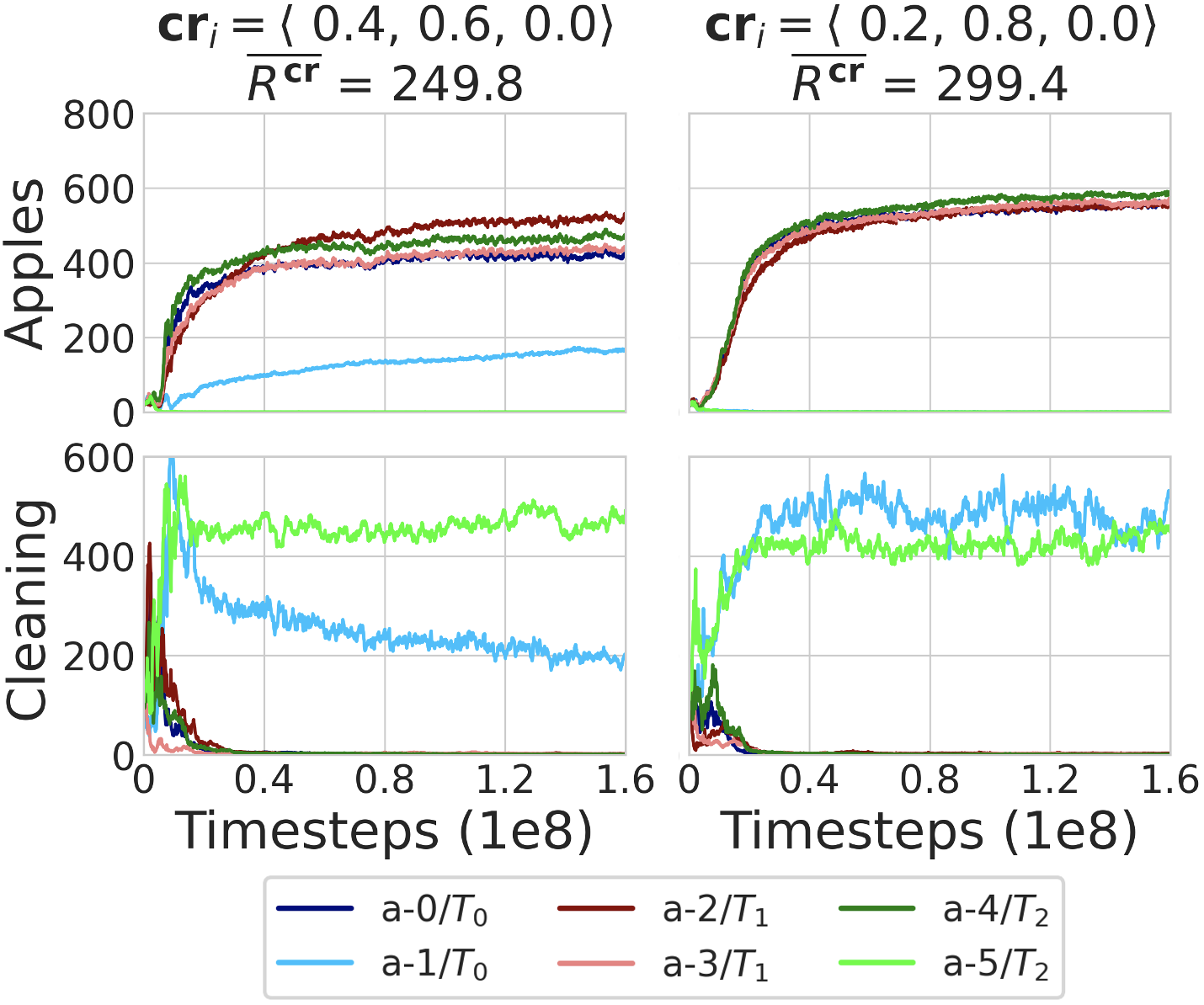}
    \caption{\textit{Cleanup:} (Scenario 2) high team-focus achieves high rewards despite a some self-focus. Each column of plots shows when team-focus is 0.6 (left) and 0.8 (right, a yellow star in Figure~\ref{fig:cleanup_results-reward}), offset with self-focus. As team-focus increases, two teams coordinate so at least one teammate cleans the river, leading to better division of labor which is maintained even when agents become fully team-focused.}
    \label{fig:cleanup_policy_highteam}
\end{figure}

\textbf{Scenario 2:}
We now analyze when agents have high team-focus compared to their other credo parameters, indicated by yellow stars in Figure~\ref{fig:cleanup_results-reward}.
Of the four experiments in this scenario, we choose $\mathbf{cr}_i = \langle 0.2, 0.8, 0.0 \rangle$ and compare with $\mathbf{cr}_i = \langle 0.4, 0.6, 0.0 \rangle$.
Since self- and team-focus change by one increment (of 0.2) and system-focus is zero, this is the simplest possible comparison.

The columns of Figure~\ref{fig:cleanup_policy_highteam} represent when agents increase team-focus from 0.6 (left) to 0.8 (right), with the remaining credo being self-focus.
When team-focus is 0.6 (left), only one team ($T_2$, green) learns to divide into the different roles of one river cleaner and one apple picker.
While a-0 on $T_0$ (dark blue) fully learns to pick apples, their teammate (a-1) does not fully learn the role of river cleaner.
This agent attempts to also pick apples and free ride on the cleaning of a-5.
$T_1$ does not commit either agent to clean the river, resulting in fewer than two full river cleaners overall.
This hinders population reward, since fewer than two total cleaners is unable to generate enough apples to support the remaining apple pickers.
Thus, the four main apple pickers only collect an average of just over 400 apples each for a mean population reward of $R^{\mathbf{cr}} = 249.8$.

In the right column when agents have higher team-focus ($\mathbf{cr}_i = \langle 0.2, 0.8, 0.0 \rangle$, yellow star scenario), two teams learn to divide into one river cleaner and one apple picker, ensuring two agents are always cleaning.
This produces enough apples for four pickers to collect about 600 each and both cleaners receive enough shared reward to overcome the incentive to free ride.
As a result, the population earns $R^{\mathbf{cr}} = 299.4$, which is 20\% more reward than when $\mathbf{cr}_i = \langle 0.4, 0.6, 0.0 \rangle$ (left) and 30\% more reward than the full common interest setting $\mathbf{cr}_i = \langle 0.0, 0.0, 1.0 \rangle$ (Figure~\ref{fig:cleanup_policy_sys-someself} left).
This division of labor is consistently learned when team-focus is high (yellow stars).

Overall, our results show specific combinations of credo support globally beneficial behavior among a population of teams.
We expand a result from~\cite{Durugkar2020BalancingIP} to social dilemmas showing some selfishness improves group performance.
Furthermore, we identify that agent specialization within their component teams results in high team-focus achieving more reward than fully system-focused credo.



\begin{figure}[t]
    \centering
    \includegraphics[width=0.8\linewidth]{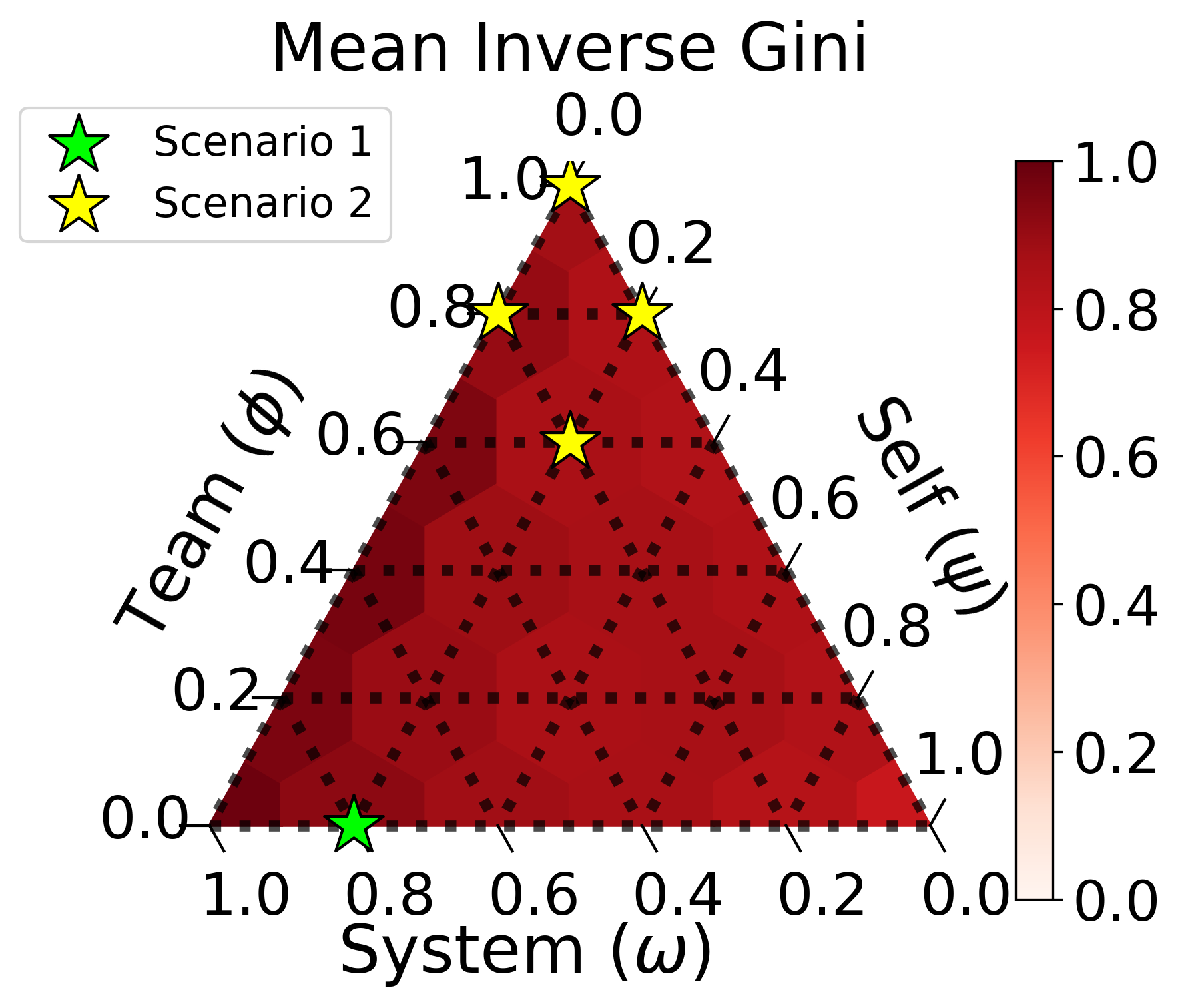}
    \caption{\textit{Cleanup:} Inverse Gini index for every credo in our evaluation. These experiments have $|\mathcal{T}| = 3$ teams of two agents each. Despite drastically different rewards, the credos which achieve high rewards also have high equality.}
    \label{fig:cleanup_results-gini}
\end{figure}

\subsubsection{Cleanup: Equality}
McKee et al.~\cite{McKee2020SocialDA} identify that analyzing an aggregated reward signal such as mean population reward may overlook settings with high reward inequality that may be detrimental to a system's stability.
Since agents do not receive any exogenous reward for cleaning the river, it is important to consider the implications and potential side effects of credo and teams on population equality, or how evenly reward is distributed among a population of agents.
We model population reward equality as the inverse Gini index, similar to past work~\cite{McKee2020SocialDA}:

\begin{equation}
    Equality = 1 - \frac{\sum_{i=0}^{N} \sum_{j=0}^{N} |R_{i}^{\mathbf{cr}} - R_{j}^{\mathbf{cr}}|}{2|N|^{2} \overline{R^{\mathbf{cr}}}},
\end{equation}

\noindent
where values closer to 1 represent more equality.
Figure~\ref{fig:cleanup_results-gini} shows our results for equality (dark red is full equality).
The full system-focused case, by definition, has perfect equality since all agents share rewards equally.
Scenario 1 still has high system-focus and also achieves high equality.
In Scenario 2, we observe the two agents that learn to clean the river always emerge from two teams learning to divide their labor.
Since each team has at least one apple picker agent, and agents have high team-focus to share rewards with cleaners, the population maintains high equality.
Both scenarios achieve more equality than fully self-focused agents $\mathbf{cr}_i = \langle 1.0, 0.0, 0.0 \rangle$ while obtaining significantly higher reward.

\section{Discussion and Future Work}
\label{sec:discussion}







We proposed a model of credo that regulates how agents optimize their behavior for different groups they belong to (i.e., self, teams, or system).
Our analysis serves as a proof of concept for exploring how agents simultaneously optimize for multiple objectives and learn to cooperate and coordinate.
Our main contributions are two-fold.
First, we show how agents are able to form cooperative policies that are robust to some amount of self-focus.
Second, we find that agents achieve high population reward with high team-focus compared to other credo parameters.
Teams are also not required to have full common interest to achieve high reward, unlike previous definitions of teams~\cite{Jaderberg2019HumanlevelPI,Baker2020EmergentTU,chung2021map}.
This has significant implications in settings where the amount of agents' self-focus may be unknown, or full alignment of a group's interests may not be possible or desired.

A key takeaway of this work is that full cooperation in a system of individual learners may not achieve the highest reward, despite several recent studies using this scenario as a basis for comparison~\cite{Wang2019EvolvingIM,gemp2020d3c,McKee2020SocialDA}.
This is contrary to the spirit of Gretzky's observation mentioned in the introduction: a certain amount of personal striving is beneficial for the system overall.
We observed that fully cooperative systems did not take full advantage of the efficiencies of labor division and task specialization, which did arise when agents had some self- or team-focus.
This tended to be particularly problematic as the number of agents sharing rewards increased, indicating a correlation between team structure, credo, and learning role-specialization.
Similar to a recent single-agent finding~\cite{arumugam2021information}, we hypothesize this is due to increased complexity in credit assignment as the reward-sharing group increases in size.

It might be possible to mitigate credit-assignment issues by actively tuning credo to regulate the amount of reward shared with teammates or the overall system.
If credo is tuned to emphasize self-focus, agents may gain a stronger direct feedback signal from their actions, guiding them towards better policies.
On the other hand, fully self-focused agents may become more cooperative if their credo is adjusted to encompass teams or the system as a whole.
In general, this may be viewed as a form of meta-learning where some credo-regulating policy shapes the environment in which agents learn.
Tuning credo could be approached in two ways: centralized or decentralized.
Using a centralized credo-tuner has the advantages of striving for global goals such as egalitarian or utilitarian ideas of equity, diversity, or productivity, though at a cost of potentially significant overhead.
On the other hand, a fully decentralized credo-tuning model may make analysing systems and the resulting equilibria even more challenging.

While past MARL work without teams tries to prevent specialization~\cite{McKee2020SocialDA,gemp2020d3c}, the areas of Team Forming and Coalition Structure Generation often construct teams specifically with specialized agents to fill roles~\cite{Andrejczyk2017Concise}.
Successful human teams, such as in sports, are specifically composed of players with diverse abilities~\cite{Radke2021Passing}.
The introduction of teams should change how specialization is viewed in MARL settings, such as agents learning emergent roles from only their credo-based reward.
If rewards can not be accessed, such as with hybrid AI/human teams, performance incentives may alter the inherent goals of agents.
Furthermore, credo may help us understand which features of a system lead to inequality, create underlying incentives, or how to achieve alternative societal goals.

We see many interesting directions for this work.
For example, analyzing situations when teams or teammates have different credo or if agents are able to dynamically tune or learn credo over time as mentioned above.
Another possible direction for future work is studying how teammates could influence the credo of their teammates or other teams and which processes achieve the highest behavioural influence.
We hope that this work inspires future directions studying multiagent teams, multi-objective optimization, and incentives to improve system performance.

\begin{acks}
This research is funded by the Natural Sciences and Engineering Research Council of Canada (NSERC), an Ontario Graduate Scholarship, a Cheriton Scholarship, and the University of Waterloo President's Graduate Scholarship.
We thank the Vector Institute for providing the compute resources necessary for this research to be conducted.
We also thank Alexi Orchard, Sriram Subramanian, Valerie Platsko, Kanav Mehra, and all reviewers for their feedback and useful discussion on earlier drafts of this work.
\end{acks}



\bibliographystyle{ACM-Reference-Format}
\bibliography{references}


\begin{thebibliography}{64}


\ifx \showCODEN    \undefined \def \showCODEN     #1{\unskip}     \fi
\ifx \showDOI      \undefined \def \showDOI       #1{#1}\fi
\ifx \showISBNx    \undefined \def \showISBNx     #1{\unskip}     \fi
\ifx \showISBNxiii \undefined \def \showISBNxiii  #1{\unskip}     \fi
\ifx \showISSN     \undefined \def \showISSN      #1{\unskip}     \fi
\ifx \showLCCN     \undefined \def \showLCCN      #1{\unskip}     \fi
\ifx \shownote     \undefined \def \shownote      #1{#1}          \fi
\ifx \showarticletitle \undefined \def \showarticletitle #1{#1}   \fi
\ifx \showURL      \undefined \def \showURL       {\relax}        \fi
\providecommand\bibfield[2]{#2}
\providecommand\bibinfo[2]{#2}
\providecommand\natexlab[1]{#1}
\providecommand\showeprint[2][]{arXiv:#2}

\bibitem[\protect\citeauthoryear{Anastassacos, Garcia, Hailes, and
  Musolesi}{Anastassacos et~al\mbox{.}}{2021}]%
        {Anastassacos2021CooperationAR}
\bibfield{author}{\bibinfo{person}{Nicolas Anastassacos},
  \bibinfo{person}{Julian Garcia}, \bibinfo{person}{S. Hailes}, {and}
  \bibinfo{person}{Mirco Musolesi}.} \bibinfo{year}{2021}\natexlab{}.
\newblock \showarticletitle{Cooperation and Reputation Dynamics with
  Reinforcement Learning}. In \bibinfo{booktitle}{\emph{AAMAS}}.
\newblock


\bibitem[\protect\citeauthoryear{Anastassacos, Hailes, and
  Musolesi}{Anastassacos et~al\mbox{.}}{2020}]%
        {Anastassacos2020PartnerSF}
\bibfield{author}{\bibinfo{person}{Nicolas Anastassacos}, \bibinfo{person}{S.
  Hailes}, {and} \bibinfo{person}{Mirco Musolesi}.}
  \bibinfo{year}{2020}\natexlab{}.
\newblock \showarticletitle{Partner Selection for the Emergence of Cooperation
  in Multi-Agent Systems Using Reinforcement Learning}. In
  \bibinfo{booktitle}{\emph{AAAI}}.
\newblock


\bibitem[\protect\citeauthoryear{Anderson and Franks}{Anderson and
  Franks}{2001}]%
        {Anderson2001TeamsIA}
\bibfield{author}{\bibinfo{person}{C. Anderson} {and} \bibinfo{person}{N.
  Franks}.} \bibinfo{year}{2001}\natexlab{}.
\newblock \showarticletitle{Teams in animal societies}.
\newblock \bibinfo{journal}{\emph{Behavioral Ecology}}  \bibinfo{volume}{12}
  (\bibinfo{year}{2001}), \bibinfo{pages}{534--540}.
\newblock


\bibitem[\protect\citeauthoryear{Anderson and Franks}{Anderson and
  Franks}{2003}]%
        {Anderson2003TeamworkIA}
\bibfield{author}{\bibinfo{person}{C. Anderson} {and} \bibinfo{person}{N.
  Franks}.} \bibinfo{year}{2003}\natexlab{}.
\newblock \showarticletitle{Teamwork in Animals, Robots, and Humans}.
\newblock \bibinfo{journal}{\emph{Advances in The Study of Behavior}}
  \bibinfo{volume}{33} (\bibinfo{year}{2003}), \bibinfo{pages}{1--48}.
\newblock


\bibitem[\protect\citeauthoryear{Andrejczuk, Rodriguez-Aguilar, and
  Sierra}{Andrejczuk et~al\mbox{.}}{2016}]%
        {Andrejczyk2017Concise}
\bibfield{author}{\bibinfo{person}{E. Andrejczuk}, \bibinfo{person}{J.~A.
  Rodriguez-Aguilar}, {and} \bibinfo{person}{C. Sierra}.}
  \bibinfo{year}{2016}\natexlab{}.
\newblock \showarticletitle{A concise review on multiagent teams: contributions
  and research opportunities}.
\newblock \bibinfo{journal}{\emph{Multi-Agent Systems and Agreement
  Technologies}} (\bibinfo{year}{2016}).
\newblock


\bibitem[\protect\citeauthoryear{Arumugam, Henderson, and Bacon}{Arumugam
  et~al\mbox{.}}{2020}]%
        {arumugam2021information}
\bibfield{author}{\bibinfo{person}{Dilip Arumugam}, \bibinfo{person}{Peter
  Henderson}, {and} \bibinfo{person}{Pierre-Luc Bacon}.}
  \bibinfo{year}{2020}\natexlab{}.
\newblock \showarticletitle{An information-theoretic perspective on credit
  assignment in reinforcement learning}.
\newblock \bibinfo{journal}{\emph{Workshop on Biological and Artificial
  Reinforcement Learning (NeurIPS 2020)}} (\bibinfo{year}{2020}).
\newblock


\bibitem[\protect\citeauthoryear{Baker}{Baker}{2020}]%
        {Baker2020EmergentRA}
\bibfield{author}{\bibinfo{person}{Bowen Baker}.}
  \bibinfo{year}{2020}\natexlab{}.
\newblock \showarticletitle{Emergent Reciprocity and Team Formation from
  Randomized Uncertain Social Preferences}.
\newblock \bibinfo{journal}{\emph{NeurIPS}} (\bibinfo{year}{2020}).
\newblock


\bibitem[\protect\citeauthoryear{Baker, Kanitscheider, Markov, Wu, Powell,
  McGrew, and Mordatch}{Baker et~al\mbox{.}}{2019}]%
        {Baker2020EmergentTU}
\bibfield{author}{\bibinfo{person}{B. Baker}, \bibinfo{person}{I.
  Kanitscheider}, \bibinfo{person}{T. Markov}, \bibinfo{person}{Y. Wu},
  \bibinfo{person}{G. Powell}, \bibinfo{person}{B. McGrew}, {and}
  \bibinfo{person}{I. Mordatch}.} \bibinfo{year}{2019}\natexlab{}.
\newblock \showarticletitle{Emergent Tool Use From Multi-Agent Autocurricula}.
  In \bibinfo{booktitle}{\emph{ICLR}}.
\newblock


\bibitem[\protect\citeauthoryear{Berner, Brockman, Chan, Cheung, Debiak,
  Dennison, Farhi, Fischer, Hashme, Hesse, J{\'o}zefowicz, Gray, Olsson,
  Pachocki, Petrov, de~Oliveira~Pinto, Raiman, Salimans, Schlatter, Schneider,
  Sidor, Sutskever, Tang, Wolski, and Zhang}{Berner et~al\mbox{.}}{2019}]%
        {Berner2019Dota2W}
\bibfield{author}{\bibinfo{person}{Christopher Berner}, \bibinfo{person}{Greg
  Brockman}, \bibinfo{person}{Brooke Chan}, \bibinfo{person}{Vicki Cheung},
  \bibinfo{person}{Przemyslaw Debiak}, \bibinfo{person}{Christy Dennison},
  \bibinfo{person}{David Farhi}, \bibinfo{person}{Quirin Fischer},
  \bibinfo{person}{Shariq Hashme}, \bibinfo{person}{Christopher Hesse},
  \bibinfo{person}{R. J{\'o}zefowicz}, \bibinfo{person}{Scott Gray},
  \bibinfo{person}{Catherine Olsson}, \bibinfo{person}{Jakub~W. Pachocki},
  \bibinfo{person}{Michael Petrov}, \bibinfo{person}{Henrique~Pond{\'e} de
  Oliveira~Pinto}, \bibinfo{person}{Jonathan Raiman}, \bibinfo{person}{Tim
  Salimans}, \bibinfo{person}{Jeremy Schlatter}, \bibinfo{person}{J.
  Schneider}, \bibinfo{person}{S. Sidor}, \bibinfo{person}{Ilya Sutskever},
  \bibinfo{person}{Jie Tang}, \bibinfo{person}{F. Wolski}, {and}
  \bibinfo{person}{Susan Zhang}.} \bibinfo{year}{2019}\natexlab{}.
\newblock \showarticletitle{Dota 2 with Large Scale Deep Reinforcement
  Learning}.
\newblock \bibinfo{journal}{\emph{ArXiv}}  \bibinfo{volume}{abs/1912.06680}
  (\bibinfo{year}{2019}).
\newblock


\bibitem[\protect\citeauthoryear{Brady}{Brady}{1993}]%
        {Brady1993GoverningTC}
\bibfield{author}{\bibinfo{person}{G. Brady}.} \bibinfo{year}{1993}\natexlab{}.
\newblock \showarticletitle{Governing the Commons: The Evolution of
  Institutions for Collective Action}.
\newblock \bibinfo{journal}{\emph{Southern Economic Journal}}
  \bibinfo{volume}{60} (\bibinfo{year}{1993}), \bibinfo{pages}{249--251}.
\newblock


\bibitem[\protect\citeauthoryear{Carter, Asencio, Trainer, DeChurch, Kanfer,
  and Zaccaro}{Carter et~al\mbox{.}}{2019}]%
        {Carter2019BestPF}
\bibfield{author}{\bibinfo{person}{D.~R. Carter}, \bibinfo{person}{R. Asencio},
  \bibinfo{person}{Hayley~M. Trainer}, \bibinfo{person}{Leslie~A. DeChurch},
  \bibinfo{person}{R. Kanfer}, {and} \bibinfo{person}{S. Zaccaro}.}
  \bibinfo{year}{2019}\natexlab{}.
\newblock \showarticletitle{Best Practices for Researchers Working in Multiteam
  Systems}.
\newblock


\bibitem[\protect\citeauthoryear{Challet and Zhang}{Challet and Zhang}{1997}]%
        {Challet1997EmergenceOC}
\bibfield{author}{\bibinfo{person}{D. Challet} {and} \bibinfo{person}{Yi-Cheng
  Zhang}.} \bibinfo{year}{1997}\natexlab{}.
\newblock \showarticletitle{Emergence of cooperation and organization in an
  evolutionary game}.
\newblock \bibinfo{journal}{\emph{Physica A-statistical Mechanics and Its
  Applications}}  \bibinfo{volume}{246} (\bibinfo{year}{1997}),
  \bibinfo{pages}{407--418}.
\newblock


\bibitem[\protect\citeauthoryear{Chung}{Chung}{2021}]%
        {chung2021map}
\bibfield{author}{\bibinfo{person}{Stephen Chung}.}
  \bibinfo{year}{2021}\natexlab{}.
\newblock \showarticletitle{MAP Propagation Algorithm: Faster Learning with a
  Team of Reinforcement Learning Agents}.
\newblock \bibinfo{journal}{\emph{Advances in Neural Information Processing
  Systems}}  \bibinfo{volume}{34} (\bibinfo{year}{2021}).
\newblock


\bibitem[\protect\citeauthoryear{Dafoe, Bachrach, Hadfield, Horvitz, Larson,
  and Graepel}{Dafoe et~al\mbox{.}}{2021}]%
        {DafoeNature2021}
\bibfield{author}{\bibinfo{person}{Allan Dafoe}, \bibinfo{person}{Yoram
  Bachrach}, \bibinfo{person}{Gillian Hadfield}, \bibinfo{person}{Eric
  Horvitz}, \bibinfo{person}{Kate Larson}, {and} \bibinfo{person}{Thore
  Graepel}.} \bibinfo{year}{2021}\natexlab{}.
\newblock \showarticletitle{Cooperative {AI}: machines must learn to find
  common ground}.
\newblock \bibinfo{journal}{\emph{Nature}}  \bibinfo{volume}{593}
  (\bibinfo{year}{2021}), \bibinfo{pages}{33--36}.
\newblock


\bibitem[\protect\citeauthoryear{Dafoe, Hughes, Bachrach, Collins, McKee,
  Leibo, Larson, and Graepel}{Dafoe et~al\mbox{.}}{2020}]%
        {Dafoe2020OpenPI}
\bibfield{author}{\bibinfo{person}{A. Dafoe}, \bibinfo{person}{Edward Hughes},
  \bibinfo{person}{Yoram Bachrach}, \bibinfo{person}{Tantum Collins},
  \bibinfo{person}{Kevin~R. McKee}, \bibinfo{person}{Joel~Z. Leibo},
  \bibinfo{person}{K. Larson}, {and} \bibinfo{person}{T. Graepel}.}
  \bibinfo{year}{2020}\natexlab{}.
\newblock \showarticletitle{Open Problems in Cooperative AI}.
\newblock \bibinfo{journal}{\emph{ArXiv}}  \bibinfo{volume}{abs/2012.08630}
  (\bibinfo{year}{2020}).
\newblock


\bibitem[\protect\citeauthoryear{Danassis, Erden, and Faltings}{Danassis
  et~al\mbox{.}}{2021}]%
        {Danassis2021ImprovedCB}
\bibfield{author}{\bibinfo{person}{Panayiotis Danassis},
  \bibinfo{person}{Zeki~Doruk Erden}, {and} \bibinfo{person}{B. Faltings}.}
  \bibinfo{year}{2021}\natexlab{}.
\newblock \showarticletitle{Improved Cooperation by Exploiting a Common
  Signal}. In \bibinfo{booktitle}{\emph{AAMAS}}.
\newblock


\bibitem[\protect\citeauthoryear{Davison, Hollenbeck, Barnes, Sleesman, and
  Ilgen}{Davison et~al\mbox{.}}{2012}]%
        {Davison2012CoordinatedAI}
\bibfield{author}{\bibinfo{person}{Robert Davison}, \bibinfo{person}{J.
  Hollenbeck}, \bibinfo{person}{Christopher~M. Barnes},
  \bibinfo{person}{Dustin~J. Sleesman}, {and} \bibinfo{person}{D.~R. Ilgen}.}
  \bibinfo{year}{2012}\natexlab{}.
\newblock \showarticletitle{Coordinated action in multiteam systems.}
\newblock \bibinfo{journal}{\emph{The Journal of applied psychology}}
  \bibinfo{volume}{97 4} (\bibinfo{year}{2012}), \bibinfo{pages}{808--24}.
\newblock


\bibitem[\protect\citeauthoryear{Dawes and Messick}{Dawes and Messick}{2000}]%
        {dawes2000}
\bibfield{author}{\bibinfo{person}{Robyn~M. Dawes} {and}
  \bibinfo{person}{David~M. Messick}.} \bibinfo{year}{2000}\natexlab{}.
\newblock \showarticletitle{Social Dilemmas}.
\newblock \bibinfo{journal}{\emph{International Journal of Psychology}}
  \bibinfo{volume}{35}, \bibinfo{number}{2} (\bibinfo{year}{2000}),
  \bibinfo{pages}{111--116}.
\newblock


\bibitem[\protect\citeauthoryear{DeChurch and Zaccaro}{DeChurch and
  Zaccaro}{2010}]%
        {DeChurch2010PerspectivesTW}
\bibfield{author}{\bibinfo{person}{Leslie~A. DeChurch} {and}
  \bibinfo{person}{S. Zaccaro}.} \bibinfo{year}{2010}\natexlab{}.
\newblock \showarticletitle{Perspectives: Teams Won’t Solve This Problem}.
\newblock \bibinfo{journal}{\emph{Human Factors: The Journal of Human Factors
  and Ergonomics Society}}  \bibinfo{volume}{52} (\bibinfo{year}{2010}),
  \bibinfo{pages}{329 -- 334}.
\newblock


\bibitem[\protect\citeauthoryear{Durugkar, Liebman, and Stone}{Durugkar
  et~al\mbox{.}}{2020}]%
        {Durugkar2020BalancingIP}
\bibfield{author}{\bibinfo{person}{Ishan Durugkar}, \bibinfo{person}{E.
  Liebman}, {and} \bibinfo{person}{P. Stone}.} \bibinfo{year}{2020}\natexlab{}.
\newblock \showarticletitle{Balancing Individual Preferences and Shared
  Objectives in Multiagent Reinforcement Learning}. In
  \bibinfo{booktitle}{\emph{IJCAI}}.
\newblock


\bibitem[\protect\citeauthoryear{Fehr and Schurtenberger}{Fehr and
  Schurtenberger}{2018}]%
        {Fehr2018NormativeFO}
\bibfield{author}{\bibinfo{person}{E. Fehr} {and} \bibinfo{person}{Ivo
  Schurtenberger}.} \bibinfo{year}{2018}\natexlab{}.
\newblock \showarticletitle{Normative foundations of human cooperation}.
\newblock \bibinfo{journal}{\emph{Nature Human Behaviour}}  \bibinfo{volume}{2}
  (\bibinfo{year}{2018}), \bibinfo{pages}{458--468}.
\newblock


\bibitem[\protect\citeauthoryear{Gemp, McKee, Everett,
  Du{\'e}{\~n}ez-Guzm{\'a}n, Bachrach, Balduzzi, and Tacchetti}{Gemp
  et~al\mbox{.}}{2022}]%
        {gemp2020d3c}
\bibfield{author}{\bibinfo{person}{Ian Gemp}, \bibinfo{person}{Kevin~R McKee},
  \bibinfo{person}{Richard Everett}, \bibinfo{person}{Edgar~A
  Du{\'e}{\~n}ez-Guzm{\'a}n}, \bibinfo{person}{Yoram Bachrach},
  \bibinfo{person}{David Balduzzi}, {and} \bibinfo{person}{Andrea Tacchetti}.}
  \bibinfo{year}{2022}\natexlab{}.
\newblock \showarticletitle{D3C: Reducing the Price of Anarchy in Multi-Agent
  Learning}.
\newblock \bibinfo{journal}{\emph{Proceedings of the 21st International
  Conference on Autonomous Agents and MultiAgent Systems}}
  (\bibinfo{year}{2022}).
\newblock


\bibitem[\protect\citeauthoryear{Grosz and Kraus}{Grosz and Kraus}{1996}]%
        {Grosz1996CollaborativePF}
\bibfield{author}{\bibinfo{person}{B. Grosz} {and} \bibinfo{person}{S. Kraus}.}
  \bibinfo{year}{1996}\natexlab{}.
\newblock \showarticletitle{Collaborative Plans for Complex Group Action}.
\newblock \bibinfo{journal}{\emph{Artif. Intell.}}  \bibinfo{volume}{86}
  (\bibinfo{year}{1996}), \bibinfo{pages}{269--357}.
\newblock


\bibitem[\protect\citeauthoryear{Grosz and Sidner}{Grosz and Sidner}{1990}]%
        {Grosz1988PlansFD}
\bibfield{author}{\bibinfo{person}{B.~J. Grosz} {and} \bibinfo{person}{C.~L.
  Sidner}.} \bibinfo{year}{1990}\natexlab{}.
\newblock \showarticletitle{Plans for Discourse}.
\newblock In \bibinfo{booktitle}{\emph{Intentions in Communication}},
  \bibfield{editor}{\bibinfo{person}{P.~R. Cohen}, \bibinfo{person}{J.~Morgan},
  {and} \bibinfo{person}{M.~E. Pollack}} (Eds.). \bibinfo{publisher}{MIT
  Press}, \bibinfo{address}{Cambridge, MA}, \bibinfo{pages}{417--444}.
\newblock


\bibitem[\protect\citeauthoryear{Herrmann, Call, Hern{\'a}ndez-Lloreda, Hare,
  and Tomasello}{Herrmann et~al\mbox{.}}{2007}]%
        {Herrmann2007HumansHE}
\bibfield{author}{\bibinfo{person}{E. Herrmann}, \bibinfo{person}{J. Call},
  \bibinfo{person}{M.~V. Hern{\'a}ndez-Lloreda}, \bibinfo{person}{B. Hare},
  {and} \bibinfo{person}{M. Tomasello}.} \bibinfo{year}{2007}\natexlab{}.
\newblock \showarticletitle{Humans Have Evolved Specialized Skills of Social
  Cognition: The Cultural Intelligence Hypothesis}.
\newblock \bibinfo{journal}{\emph{Science}}  \bibinfo{volume}{317}
  (\bibinfo{year}{2007}), \bibinfo{pages}{1360 -- 1366}.
\newblock


\bibitem[\protect\citeauthoryear{Hostallero, Kim, Moon, Son, Kang, and
  Yi}{Hostallero et~al\mbox{.}}{2020}]%
        {hostallero2020inducing}
\bibfield{author}{\bibinfo{person}{David~Earl Hostallero},
  \bibinfo{person}{Daewoo Kim}, \bibinfo{person}{Sangwoo Moon},
  \bibinfo{person}{Kyunghwan Son}, \bibinfo{person}{Wan~Ju Kang}, {and}
  \bibinfo{person}{Yung Yi}.} \bibinfo{year}{2020}\natexlab{}.
\newblock \showarticletitle{Inducing cooperation through reward reshaping based
  on peer evaluations in deep multi-agent reinforcement learning}. In
  \bibinfo{booktitle}{\emph{Proceedings of the 19th International Conference on
  Autonomous Agents and MultiAgent Systems}}. \bibinfo{pages}{520--528}.
\newblock


\bibitem[\protect\citeauthoryear{Hughes, Leibo, Phillips, Tuyls,
  Du{\'e}{\~n}ez-Guzm{\'a}n, Casta{\~n}eda, Dunning, Zhu, McKee, Koster, Roff,
  and Graepel}{Hughes et~al\mbox{.}}{2018}]%
        {Hughes2018InequityAI}
\bibfield{author}{\bibinfo{person}{Edward Hughes}, \bibinfo{person}{Joel~Z.
  Leibo}, \bibinfo{person}{Matthew Phillips}, \bibinfo{person}{K. Tuyls},
  \bibinfo{person}{Edgar~A. Du{\'e}{\~n}ez-Guzm{\'a}n}, \bibinfo{person}{A.
  Casta{\~n}eda}, \bibinfo{person}{Iain Dunning}, \bibinfo{person}{Tina Zhu},
  \bibinfo{person}{Kevin~R. McKee}, \bibinfo{person}{R. Koster},
  \bibinfo{person}{Heather Roff}, {and} \bibinfo{person}{T. Graepel}.}
  \bibinfo{year}{2018}\natexlab{}.
\newblock \showarticletitle{Inequity aversion improves cooperation in
  intertemporal social dilemmas}. In \bibinfo{booktitle}{\emph{NeurIPS}}.
\newblock


\bibitem[\protect\citeauthoryear{Jaderberg, Czarnecki, Dunning, Marris, Lever,
  Casta{\~n}eda, Beattie, Rabinowitz, Morcos, Ruderman, Sonnerat, Green,
  Deason, Leibo, Silver, Hassabis, Kavukcuoglu, and Graepel}{Jaderberg
  et~al\mbox{.}}{2019}]%
        {Jaderberg2019HumanlevelPI}
\bibfield{author}{\bibinfo{person}{Max Jaderberg}, \bibinfo{person}{Wojciech
  Czarnecki}, \bibinfo{person}{Iain Dunning}, \bibinfo{person}{Luke Marris},
  \bibinfo{person}{Guy Lever}, \bibinfo{person}{A. Casta{\~n}eda},
  \bibinfo{person}{Charlie Beattie}, \bibinfo{person}{Neil~C. Rabinowitz},
  \bibinfo{person}{Ari~S. Morcos}, \bibinfo{person}{Avraham Ruderman},
  \bibinfo{person}{Nicolas Sonnerat}, \bibinfo{person}{Tim Green},
  \bibinfo{person}{Louise Deason}, \bibinfo{person}{Joel~Z. Leibo},
  \bibinfo{person}{D. Silver}, \bibinfo{person}{D. Hassabis},
  \bibinfo{person}{K. Kavukcuoglu}, {and} \bibinfo{person}{T. Graepel}.}
  \bibinfo{year}{2019}\natexlab{}.
\newblock \showarticletitle{Human-level performance in 3D multiplayer games
  with population-based reinforcement learning}.
\newblock \bibinfo{journal}{\emph{Science}}  \bibinfo{volume}{364}
  (\bibinfo{year}{2019}), \bibinfo{pages}{859 -- 865}.
\newblock


\bibitem[\protect\citeauthoryear{Jaques, Lazaridou, Hughes, Çaglar
  G{\"u}lçehre, Ortega, Strouse, Leibo, and Freitas}{Jaques
  et~al\mbox{.}}{2019}]%
        {Jaques2019SocialIA}
\bibfield{author}{\bibinfo{person}{Natasha Jaques}, \bibinfo{person}{A.
  Lazaridou}, \bibinfo{person}{Edward Hughes}, \bibinfo{person}{Çaglar
  G{\"u}lçehre}, \bibinfo{person}{Pedro~A. Ortega}, \bibinfo{person}{D.
  Strouse}, \bibinfo{person}{Joel~Z. Leibo}, {and} \bibinfo{person}{N.~D.
  Freitas}.} \bibinfo{year}{2019}\natexlab{}.
\newblock \showarticletitle{Social Influence as Intrinsic Motivation for
  Multi-Agent Deep Reinforcement Learning}. In
  \bibinfo{booktitle}{\emph{ICML}}.
\newblock


\bibitem[\protect\citeauthoryear{Kitano, Asada, Kuniyoshi, Noda, and
  Osawa}{Kitano et~al\mbox{.}}{1997}]%
        {Kitano1997RoboCupTR}
\bibfield{author}{\bibinfo{person}{H. Kitano}, \bibinfo{person}{M. Asada},
  \bibinfo{person}{Y. Kuniyoshi}, \bibinfo{person}{I. Noda}, {and}
  \bibinfo{person}{Eiichi Osawa}.} \bibinfo{year}{1997}\natexlab{}.
\newblock \showarticletitle{RoboCup: The Robot World Cup Initiative}. In
  \bibinfo{booktitle}{\emph{AGENTS '97}}.
\newblock


\bibitem[\protect\citeauthoryear{Leibo, Zambaldi, Lanctot, Marecki, and
  Graepel}{Leibo et~al\mbox{.}}{2017}]%
        {Leibo2017MultiagentRL}
\bibfield{author}{\bibinfo{person}{J.~Z. Leibo}, \bibinfo{person}{V. Zambaldi},
  \bibinfo{person}{M. Lanctot}, \bibinfo{person}{J. Marecki}, {and}
  \bibinfo{person}{T. Graepel}.} \bibinfo{year}{2017}\natexlab{}.
\newblock \showarticletitle{Multi-agent Reinforcement Learning in Sequential
  Social Dilemmas}. In \bibinfo{booktitle}{\emph{AAMAS-17}}.
\newblock


\bibitem[\protect\citeauthoryear{Luciano, DeChurch, and Mathieu}{Luciano
  et~al\mbox{.}}{2018}]%
        {Luciano2018MultiteamSA}
\bibfield{author}{\bibinfo{person}{Margaret~M Luciano},
  \bibinfo{person}{Leslie~A. DeChurch}, {and} \bibinfo{person}{J. Mathieu}.}
  \bibinfo{year}{2018}\natexlab{}.
\newblock \showarticletitle{Multiteam Systems: A Structural Framework and
  Meso-Theory of System Functioning}.
\newblock \bibinfo{journal}{\emph{Journal of Management}}  \bibinfo{volume}{44}
  (\bibinfo{year}{2018}), \bibinfo{pages}{1065 -- 1096}.
\newblock


\bibitem[\protect\citeauthoryear{Macke, Mirsky, and Stone}{Macke
  et~al\mbox{.}}{2021}]%
        {Macke2021ExpectedVO}
\bibfield{author}{\bibinfo{person}{W. Macke}, \bibinfo{person}{R. Mirsky},
  {and} \bibinfo{person}{P. Stone}.} \bibinfo{year}{2021}\natexlab{}.
\newblock \showarticletitle{Expected Value of Communication for Planning in Ad
  Hoc Teamwork}. In \bibinfo{booktitle}{\emph{AAAI-21}}.
\newblock


\bibitem[\protect\citeauthoryear{Mathieu, Marks, and Zaccaro}{Mathieu
  et~al\mbox{.}}{2001}]%
        {Mathieu2001MTSs}
\bibfield{author}{\bibinfo{person}{J.~E. Mathieu}, \bibinfo{person}{M.~A.
  Marks}, {and} \bibinfo{person}{S.~J. Zaccaro}.}
  \bibinfo{year}{2001}\natexlab{}.
\newblock \showarticletitle{Multiteam Systems}.
\newblock \bibinfo{journal}{\emph{Handbook of Industrial, Work, and
  Ordanizational Psychology}}  \bibinfo{volume}{2} (\bibinfo{year}{2001}).
\newblock


\bibitem[\protect\citeauthoryear{McKee, Gemp, McWilliams,
  Du{\'e}{\~n}ez-Guzm{\'a}n, Hughes, and Leibo}{McKee et~al\mbox{.}}{2020}]%
        {McKee2020SocialDA}
\bibfield{author}{\bibinfo{person}{Kevin~R. McKee}, \bibinfo{person}{I. Gemp},
  \bibinfo{person}{Brian McWilliams}, \bibinfo{person}{Edgar~A.
  Du{\'e}{\~n}ez-Guzm{\'a}n}, \bibinfo{person}{Edward Hughes}, {and}
  \bibinfo{person}{Joel~Z. Leibo}.} \bibinfo{year}{2020}\natexlab{}.
\newblock \showarticletitle{Social Diversity and Social Preferences in
  Mixed-Motive Reinforcement Learning}.
\newblock \bibinfo{journal}{\emph{AAMAS}} (\bibinfo{year}{2020}).
\newblock


\bibitem[\protect\citeauthoryear{Mirsky, Macke, Wang, Yedidsion, and
  Stone}{Mirsky et~al\mbox{.}}{2020}]%
        {Mirsky2020APF}
\bibfield{author}{\bibinfo{person}{Reuth Mirsky}, \bibinfo{person}{William
  Macke}, \bibinfo{person}{A. Wang}, \bibinfo{person}{Harel Yedidsion}, {and}
  \bibinfo{person}{P. Stone}.} \bibinfo{year}{2020}\natexlab{}.
\newblock \showarticletitle{A Penny for Your Thoughts: The Value of
  Communication in Ad Hoc Teamwork}. In \bibinfo{booktitle}{\emph{IJCAI}}.
\newblock


\bibitem[\protect\citeauthoryear{Mnih, Kavukcuoglu, Silver, Rusu, Veness,
  Bellemare, Graves, Riedmiller, Fidjeland, Ostrovski, Petersen, Beattie,
  Sadik, Antonoglou, King, Kumaran, Wierstra, Legg, and Hassabis}{Mnih
  et~al\mbox{.}}{2015}]%
        {Mnih2015HumanlevelCT}
\bibfield{author}{\bibinfo{person}{V. Mnih}, \bibinfo{person}{K. Kavukcuoglu},
  \bibinfo{person}{D. Silver}, \bibinfo{person}{A.~A. Rusu},
  \bibinfo{person}{J. Veness}, \bibinfo{person}{M.~G. Bellemare},
  \bibinfo{person}{A. Graves}, \bibinfo{person}{M.~A. Riedmiller},
  \bibinfo{person}{A. Fidjeland}, \bibinfo{person}{G. Ostrovski},
  \bibinfo{person}{S. Petersen}, \bibinfo{person}{C. Beattie},
  \bibinfo{person}{A. Sadik}, \bibinfo{person}{I. Antonoglou},
  \bibinfo{person}{H. King}, \bibinfo{person}{D. Kumaran}, \bibinfo{person}{D.
  Wierstra}, \bibinfo{person}{S. Legg}, {and} \bibinfo{person}{D. Hassabis}.}
  \bibinfo{year}{2015}\natexlab{}.
\newblock \showarticletitle{Human-level control through deep reinforcement
  learning}.
\newblock \bibinfo{journal}{\emph{Nature}}  \bibinfo{volume}{518}
  (\bibinfo{year}{2015}), \bibinfo{pages}{529--533}.
\newblock


\bibitem[\protect\citeauthoryear{Ostrom}{Ostrom}{1990}]%
        {Ostrom1990GoverningTC}
\bibfield{author}{\bibinfo{person}{E. Ostrom}.}
  \bibinfo{year}{1990}\natexlab{}.
\newblock \showarticletitle{Governing the Commons: The Evolution of
  Institutions for Collective Action}.
\newblock \bibinfo{journal}{\emph{Natural Resources Journal}}
  \bibinfo{volume}{32} (\bibinfo{year}{1990}), \bibinfo{pages}{415}.
\newblock


\bibitem[\protect\citeauthoryear{Peysakhovich and Lerer}{Peysakhovich and
  Lerer}{2018}]%
        {prosocialStag2018}
\bibfield{author}{\bibinfo{person}{Alexander Peysakhovich} {and}
  \bibinfo{person}{Adam Lerer}.} \bibinfo{year}{2018}\natexlab{}.
\newblock \showarticletitle{Prosocial Learning Agents Solve Generalized Stag
  Hunts Better than Selfish Ones}. In \bibinfo{booktitle}{\emph{Proceedings of
  the 17th International Conference on Autonomous Agents and MultiAgent
  Systems}} (Stockholm, Sweden) \emph{(\bibinfo{series}{AAMAS '18})}.
  \bibinfo{publisher}{International Foundation for Autonomous Agents and
  Multiagent Systems}, \bibinfo{address}{Richland, SC},
  \bibinfo{pages}{2043–2044}.
\newblock


\bibitem[\protect\citeauthoryear{Pollack}{Pollack}{1986}]%
        {Pollack1986AMO}
\bibfield{author}{\bibinfo{person}{M. Pollack}.}
  \bibinfo{year}{1986}\natexlab{}.
\newblock \showarticletitle{A Model of Plan Inference that Distinguishes
  between the Beliefs of Actors and observers}. In
  \bibinfo{booktitle}{\emph{ACL}}.
\newblock


\bibitem[\protect\citeauthoryear{Pollack}{Pollack}{1990}]%
        {Pollack90plansas}
\bibfield{author}{\bibinfo{person}{Martha~E. Pollack}.}
  \bibinfo{year}{1990}\natexlab{}.
\newblock \showarticletitle{Plans As Complex Mental Attitudes}. In
  \bibinfo{booktitle}{\emph{Intentions in Communication}}.
  \bibinfo{publisher}{MIT Press}, \bibinfo{pages}{77--103}.
\newblock


\bibitem[\protect\citeauthoryear{Porck, Matta, Hollenbeck, Oh, Lanaj, and
  Lee}{Porck et~al\mbox{.}}{2019}]%
        {Porck2019SocialII}
\bibfield{author}{\bibinfo{person}{J. Porck}, \bibinfo{person}{Fadel~K Matta},
  \bibinfo{person}{J. Hollenbeck}, \bibinfo{person}{Jo~K. Oh},
  \bibinfo{person}{Klodiana Lanaj}, {and} \bibinfo{person}{S. Lee}.}
  \bibinfo{year}{2019}\natexlab{}.
\newblock \showarticletitle{Social Identification in Multiteam Systems: The
  Role of Depletion and Task Complexity}.
\newblock \bibinfo{journal}{\emph{Academy of Management Journal}}
  \bibinfo{volume}{62} (\bibinfo{year}{2019}), \bibinfo{pages}{1137--1162}.
\newblock


\bibitem[\protect\citeauthoryear{Radke, Larson, and Brecht}{Radke
  et~al\mbox{.}}{2022}]%
        {Radke2022Exploring}
\bibfield{author}{\bibinfo{person}{David Radke}, \bibinfo{person}{Kate Larson},
  {and} \bibinfo{person}{Tim Brecht}.} \bibinfo{year}{2022}\natexlab{}.
\newblock \showarticletitle{Exploring the Benefits of Teams in Multiagent
  Learning}. In \bibinfo{booktitle}{\emph{IJCAI}}.
\newblock


\bibitem[\protect\citeauthoryear{Radke, Radke, Brecht, and Pawelczyk}{Radke
  et~al\mbox{.}}{2021}]%
        {Radke2021Passing}
\bibfield{author}{\bibinfo{person}{D.~T. Radke}, \bibinfo{person}{D.~L. Radke},
  \bibinfo{person}{T. Brecht}, {and} \bibinfo{person}{A. Pawelczyk}.}
  \bibinfo{year}{2021}\natexlab{}.
\newblock \showarticletitle{Passing and Pressure Metrics in Ice Hockey}.
\newblock \bibinfo{journal}{\emph{Workshop of AI for Sports Analytics}}
  (\bibinfo{year}{2021}).
\newblock


\bibitem[\protect\citeauthoryear{Rand and Nowak}{Rand and Nowak}{2013}]%
        {Rand2013HumanC}
\bibfield{author}{\bibinfo{person}{David~G. Rand} {and} \bibinfo{person}{M.
  Nowak}.} \bibinfo{year}{2013}\natexlab{}.
\newblock \showarticletitle{Human cooperation}.
\newblock \bibinfo{journal}{\emph{Trends in Cognitive Sciences}}
  \bibinfo{volume}{17} (\bibinfo{year}{2013}), \bibinfo{pages}{413--425}.
\newblock


\bibitem[\protect\citeauthoryear{Rapoport}{Rapoport}{1974}]%
        {Rapoport1974PrisonersD}
\bibfield{author}{\bibinfo{person}{A. Rapoport}.}
  \bibinfo{year}{1974}\natexlab{}.
\newblock \showarticletitle{Prisoner’s Dilemma — Recollections and
  Observations}.
\newblock


\bibitem[\protect\citeauthoryear{Richerson}{Richerson}{1998}]%
        {Richerson1998EvolutionOfHuman}
\bibfield{author}{\bibinfo{person}{Peter Richerson}.}
  \bibinfo{year}{1998}\natexlab{}.
\newblock \bibinfo{booktitle}{\emph{The Evolution of Human Ultra-sociality}
  (\bibinfo{edition}{2.01} ed.)}.
\newblock \bibinfo{publisher}{University of California, Davis},
  \bibinfo{address}{Davis, California}.
\newblock


\bibitem[\protect\citeauthoryear{Ryu, Shin, and Park}{Ryu
  et~al\mbox{.}}{2020}]%
        {Ryu2020MultiAgentAW}
\bibfield{author}{\bibinfo{person}{Heechang Ryu}, \bibinfo{person}{Hayong
  Shin}, {and} \bibinfo{person}{Jinkyoo Park}.}
  \bibinfo{year}{2020}\natexlab{}.
\newblock \showarticletitle{Multi-Agent Actor-Critic with Hierarchical Graph
  Attention Network}. In \bibinfo{booktitle}{\emph{AAAI}}.
\newblock


\bibitem[\protect\citeauthoryear{Ryu, Shin, and Park}{Ryu
  et~al\mbox{.}}{2021}]%
        {Ryu2021CooperativeAC}
\bibfield{author}{\bibinfo{person}{Heechang Ryu}, \bibinfo{person}{Hayong
  Shin}, {and} \bibinfo{person}{Jinkyoo Park}.}
  \bibinfo{year}{2021}\natexlab{}.
\newblock \showarticletitle{Cooperative and Competitive Biases for Multi-Agent
  Reinforcement Learning}. In \bibinfo{booktitle}{\emph{AAMAS}}.
\newblock


\bibitem[\protect\citeauthoryear{Santos, Santos, and Pacheco}{Santos
  et~al\mbox{.}}{2018}]%
        {Santos2018SocialNC}
\bibfield{author}{\bibinfo{person}{F. Santos}, \bibinfo{person}{F.~C. Santos},
  {and} \bibinfo{person}{J. Pacheco}.} \bibinfo{year}{2018}\natexlab{}.
\newblock \showarticletitle{Social norm complexity and past reputations in the
  evolution of cooperation}.
\newblock \bibinfo{journal}{\emph{Nature}}  \bibinfo{volume}{555}
  (\bibinfo{year}{2018}), \bibinfo{pages}{242--245}.
\newblock


\bibitem[\protect\citeauthoryear{Schnell, Schimmelpfennig, and
  Muthukrishna}{Schnell et~al\mbox{.}}{2021}]%
        {Schnell2021levels}
\bibfield{author}{\bibinfo{person}{E. Schnell}, \bibinfo{person}{R.
  Schimmelpfennig}, {and} \bibinfo{person}{M. Muthukrishna}.}
  \bibinfo{year}{2021}\natexlab{}.
\newblock \showarticletitle{The Size of the Stag Determines the Level of
  Cooperation}.
\newblock \bibinfo{journal}{\emph{bioRxiv}} (\bibinfo{year}{2021}).
\newblock


\bibitem[\protect\citeauthoryear{Schulman, Wolski, Dhariwal, Radford, and
  Klimov}{Schulman et~al\mbox{.}}{2017}]%
        {PPO2017}
\bibfield{author}{\bibinfo{person}{John Schulman}, \bibinfo{person}{Filip
  Wolski}, \bibinfo{person}{Prafulla Dhariwal}, \bibinfo{person}{Alec Radford},
  {and} \bibinfo{person}{Oleg Klimov}.} \bibinfo{year}{2017}\natexlab{}.
\newblock \showarticletitle{Proximal Policy Optimization Algorithms}.
\newblock \bibinfo{journal}{\emph{CoRR}} (\bibinfo{year}{2017}).
\newblock


\bibitem[\protect\citeauthoryear{Simpson and Weiner}{Simpson and
  Weiner}{1989}]%
        {OxfordDictionary}
\bibfield{author}{\bibinfo{person}{J.~A. Simpson} {and}
  \bibinfo{person}{E.~S.~C. Weiner}.} \bibinfo{year}{1989}\natexlab{}.
\newblock \bibinfo{booktitle}{\emph{{The Oxford English Dictionary}}}.
\newblock \bibinfo{publisher}{Oxford University Press}.
\newblock


\bibitem[\protect\citeauthoryear{Stone, Kaminka, Kraus, and Rosenschein}{Stone
  et~al\mbox{.}}{2010}]%
        {Stone2010AdHA}
\bibfield{author}{\bibinfo{person}{P. Stone}, \bibinfo{person}{G. Kaminka},
  \bibinfo{person}{S. Kraus}, {and} \bibinfo{person}{J. Rosenschein}.}
  \bibinfo{year}{2010}\natexlab{}.
\newblock \showarticletitle{Ad Hoc Autonomous Agent Teams: Collaboration
  without Pre-Coordination}. In \bibinfo{booktitle}{\emph{AAAI}}.
\newblock


\bibitem[\protect\citeauthoryear{Sundstrom, Meuse, and Futrell}{Sundstrom
  et~al\mbox{.}}{1990}]%
        {Sundstrom1990WorkTA}
\bibfield{author}{\bibinfo{person}{E. Sundstrom}, \bibinfo{person}{K.~P.~D.
  Meuse}, {and} \bibinfo{person}{D. Futrell}.} \bibinfo{year}{1990}\natexlab{}.
\newblock \showarticletitle{Work teams: Applications and effectiveness.}
\newblock \bibinfo{journal}{\emph{American Psychologist}}  \bibinfo{volume}{45}
  (\bibinfo{year}{1990}), \bibinfo{pages}{120--133}.
\newblock


\bibitem[\protect\citeauthoryear{Tambe}{Tambe}{1997}]%
        {Tambe1997TowardsFT}
\bibfield{author}{\bibinfo{person}{Milind Tambe}.}
  \bibinfo{year}{1997}\natexlab{}.
\newblock \showarticletitle{Towards Flexible Teamwork}.
\newblock \bibinfo{journal}{\emph{J. Artif. Intell. Res.}}  \bibinfo{volume}{7}
  (\bibinfo{year}{1997}), \bibinfo{pages}{83--124}.
\newblock


\bibitem[\protect\citeauthoryear{Tomasello, Melis, Tennie, Wyman, and
  Herrmann}{Tomasello et~al\mbox{.}}{2012}]%
        {Tomasello2012TwoKS}
\bibfield{author}{\bibinfo{person}{M. Tomasello}, \bibinfo{person}{A. Melis},
  \bibinfo{person}{C. Tennie}, \bibinfo{person}{Emily Wyman}, {and}
  \bibinfo{person}{E. Herrmann}.} \bibinfo{year}{2012}\natexlab{}.
\newblock \showarticletitle{Two Key Steps in the Evolution of Human
  Cooperation}.
\newblock \bibinfo{journal}{\emph{Current Anthropology}}  \bibinfo{volume}{53}
  (\bibinfo{year}{2012}), \bibinfo{pages}{673 -- 692}.
\newblock


\bibitem[\protect\citeauthoryear{Tomasello and Vaish}{Tomasello and
  Vaish}{2013}]%
        {Tomasello2013OriginsOH}
\bibfield{author}{\bibinfo{person}{M. Tomasello} {and} \bibinfo{person}{Amrisha
  Vaish}.} \bibinfo{year}{2013}\natexlab{}.
\newblock \showarticletitle{Origins of human cooperation and morality.}
\newblock \bibinfo{journal}{\emph{Annual review of psychology}}
  \bibinfo{volume}{64} (\bibinfo{year}{2013}), \bibinfo{pages}{231--55}.
\newblock


\bibitem[\protect\citeauthoryear{Vinitsky, Jaques, Leibo, Castenada, and
  Hughes}{Vinitsky et~al\mbox{.}}{2019}]%
        {SSDOpenSource}
\bibfield{author}{\bibinfo{person}{Eugene Vinitsky}, \bibinfo{person}{Natasha
  Jaques}, \bibinfo{person}{Joel Leibo}, \bibinfo{person}{Antonio Castenada},
  {and} \bibinfo{person}{Edward Hughes}.} \bibinfo{year}{2019}\natexlab{}.
\newblock \bibinfo{title}{An Open Source Implementation of Sequential Social
  Dilemma Games}.
\newblock
  \bibinfo{howpublished}{\url{https://github.com/eugenevinitsky/sequential_social_dilemma_games/issues/182}}.
\newblock
\newblock
\shownote{GitHub repository.}


\bibitem[\protect\citeauthoryear{Wang, Hughes, Fernando, Czarnecki,
  Du{\'e}{\~n}ez-Guzm{\'a}n, and Leibo}{Wang et~al\mbox{.}}{2019}]%
        {Wang2019EvolvingIM}
\bibfield{author}{\bibinfo{person}{J.~X. Wang}, \bibinfo{person}{E. Hughes},
  \bibinfo{person}{C. Fernando}, \bibinfo{person}{W.~M. Czarnecki},
  \bibinfo{person}{E.~A. Du{\'e}{\~n}ez-Guzm{\'a}n}, {and}
  \bibinfo{person}{J.~Z. Leibo}.} \bibinfo{year}{2019}\natexlab{}.
\newblock \showarticletitle{Evolving Intrinsic Motivations for Altruistic
  Behavior}. In \bibinfo{booktitle}{\emph{AAMAS-19}}.
  \bibinfo{pages}{683--692}.
\newblock


\bibitem[\protect\citeauthoryear{Wijnmaalen, Voordijk, Rietjens, and
  Dewulf}{Wijnmaalen et~al\mbox{.}}{2019}]%
        {Wijnmaalen2019IntergroupBI}
\bibfield{author}{\bibinfo{person}{Julia Wijnmaalen}, \bibinfo{person}{H.
  Voordijk}, \bibinfo{person}{S. Rietjens}, {and} \bibinfo{person}{G. Dewulf}.}
  \bibinfo{year}{2019}\natexlab{}.
\newblock \showarticletitle{Intergroup behavior in military multiteam systems}.
\newblock \bibinfo{journal}{\emph{Human Relations}}  \bibinfo{volume}{72}
  (\bibinfo{year}{2019}), \bibinfo{pages}{1081 -- 1104}.
\newblock


\bibitem[\protect\citeauthoryear{Yang, Li, Farajtabar, Sunehag, Hughes, and
  Zha}{Yang et~al\mbox{.}}{2020}]%
        {yang2020learning}
\bibfield{author}{\bibinfo{person}{Jiachen Yang}, \bibinfo{person}{Ang Li},
  \bibinfo{person}{Mehrdad Farajtabar}, \bibinfo{person}{Peter Sunehag},
  \bibinfo{person}{Edward Hughes}, {and} \bibinfo{person}{Hongyuan Zha}.}
  \bibinfo{year}{2020}\natexlab{}.
\newblock \showarticletitle{Learning to incentivize other learning agents}.
\newblock \bibinfo{journal}{\emph{Advances in Neural Information Processing
  Systems}}  \bibinfo{volume}{33} (\bibinfo{year}{2020}),
  \bibinfo{pages}{15208--15219}.
\newblock


\bibitem[\protect\citeauthoryear{Zaccaro, Marks, and DeChurch}{Zaccaro
  et~al\mbox{.}}{2012}]%
        {Zaccaro2012MultiteamSA}
\bibfield{author}{\bibinfo{person}{S. Zaccaro}, \bibinfo{person}{M. Marks},
  {and} \bibinfo{person}{Leslie~A. DeChurch}.} \bibinfo{year}{2012}\natexlab{}.
\newblock \showarticletitle{Multiteam Systems: An Introduction}.
\newblock


\bibitem[\protect\citeauthoryear{Zaccaro, Dubrow, Torres, and Campbell}{Zaccaro
  et~al\mbox{.}}{2020}]%
        {Zaccaro2020MultiteamSA}
\bibfield{author}{\bibinfo{person}{Stephen~J. Zaccaro},
  \bibinfo{person}{Samantha Dubrow}, \bibinfo{person}{Elisa~M. Torres}, {and}
  \bibinfo{person}{Lauren~N.P. Campbell}.} \bibinfo{year}{2020}\natexlab{}.
\newblock \showarticletitle{Multiteam Systems: An Integrated Review and
  Comparison of Different Forms}.
\newblock \bibinfo{journal}{\emph{Annual Review of Organizational Psychology
  and Organizational Behavior}} \bibinfo{volume}{7}, \bibinfo{number}{1}
  (\bibinfo{year}{2020}), \bibinfo{pages}{479--503}.
\newblock


\end{thebibliography}


\clearpage

\appendix

\section{Environment Configurations}

\subsection{Iterated Prisoner's Dilemma (IPD)}
\label{sec:appendix_ipd}

\subsubsection{Matching Algorithm}

At each instance of the IPD, agent $i$ is given a counterpart, $j$ to play the game in Table~\ref{tab:pd_table}.
Agent pairings are assigned subject to $\inteam$, the probability that the counterpart is a teammate.
In our analysis, we experiment with three different environments with $|\mathcal{T}| = 5$ teams of five agents each.
When $\inteam = 0.06$ agents are four times more likely to receive a counterpart from another team.
When $\inteam = 0.2$, a counterpart from any of the five teams has equal likelihood.
When $\inteam = 0.5$, agents are four times more likely to receive a counterpart from their own team.
Each episode, we construct $N$ pairings by matching each agent $i\in N$ to a counterpart $j\in N\setminus\{i\}$ sampled subject to $\inteam$.
Each agent observes the team their counterpart belongs to through a numerical signal $s_i \in S$, but not their actual individual identity.

\subsubsection{IPD RL Algorithm}
\label{subsec:MARL_exp}

For each round of the IPD, agent $i$ is paired with another agent $j$ chosen randomly from the population, subject to the probability of being paired with a teammate $\inteam$.
The two agents play one iteration of the game shown in Table \ref{tab:pd_table}.
Each agent observes the team their counterpart belongs to instead their actual identity. 
In particular, for a pair of agents, $i$ and $j$, their states $s_i$ and $s_j$ are defined as   $s_i = T_j$ and $s_j = T_i$.
Given each $s$, the two agents simultaneously choose actions $a_i$ and $a_j$, either to cooperate or defect.
They do not observe the action of their counterpart, but instead receive rewards $R_i^{\mathbf{cr}}$ and $R_j^{\mathbf{cr}}$ based on every interaction and their credo.
Each $i$ (and also $j$ with their information) stores the tuple $\langle s_i, a_i, R_i^{\mathbf{cr}} \rangle$ in their replay buffer to train their policy after each episode using Deep $Q$-Learning~\cite{Mnih2015HumanlevelCT} by sampling a random batch of 32 interactions.
Each agent's internal neural network consists of an input layer of size $|\mathcal{T}| = 5$, two hidden layers of 200 nodes each with hyperbolic tangent activation functions, and a two-action output layer with a linear activation function.
Our agents use a learning rate of $1 \times 10^{-4}$ and discount factor of $0.99$ with $\epsilon$-exploration.

\begin{table}[t]
\begin{center}
 \begin{tabular}{|c||c|c|} 
 \hline
  & Cooperate & Defect \\ [0.5ex] 
 \hline\hline
Cooperate & $b-c$, $b-c$ & $-c$, $b$ \\\hline
Defect & $b$,  $-c$ & 0, 0 \\
 \hline
\end{tabular}
\caption{An example of the Prisoner's Dilemma with the costs (c) and benefits (b) of cooperating ($b>c>0$).}
\label{tab:pd_table}
\end{center}
\end{table}

\subsection{Cleanup Markov Game}
\label{sec:appendix_cleanup}

\subsubsection{Cleanup RL Algorithm}
\label{sec:cleanup_rl_alg}

The typical environmental setup for Cleanup implemented in past work uses five agents.
However, five agents are unable to create multiple teams with the same size.
Therefore, we instantiate $N = 6$ agents to create $|\mathcal{T}| = 3$ teams of two agents each.
We deploy the default Proximal Policy Optimization (PPO)~\cite{PPO2017} learning algorithm from the Cleanup repository~\cite{SSDOpenSource}.
PPO is a policy gradient algorithm which constrains the space of policy updates to avoid large policy updates for smoother training and has been previously shown as a good algorithm for agents to learn in Cleanup.
Agents are only able to observe the a $15 \times 15$ box centered at their location and update their policies using the environment's default batch size of at least 16,000 timesteps and a maximum of 100,000 timesteps. 
Each episode executes for 1,000 timesteps and we run experiments for $1.6 \times 10^8$ timesteps.
The learning rate decreases linearly from $1.2 \times 10^{-3}$ to $1.2 \times 10^{-5}$ over the first $2 \times 10^7$ timesteps and remains static afterwards.

\subsubsection{Environment Parameters}

An example of Cleanup with three teams is shown in Figure~\ref{fig:cleanup_env}.
The social dilemma dynamics of Cleanup rely on the existence of waste and apples.
The functions which govern the creation of these features can be easily modified; however, we evaluate our model of teams primarily using the default parameters of the environment which include a waste regeneration rate, waste regeneration threshold, and apple generation rate.
Once less than 40\% of the river (aquifer) grid-cells contain waste, waste regenerates at each clean cell with a probability of 50\% at each timestep.
Below this waste regeneration, apples spawn at each location in the orchard with a linear probability ranging from 0\% when waste is 40\% of the river up to 5\% when waste makes up 0\% of the river.

\begin{figure}[t]
    \centering
    \includegraphics[width=\linewidth]{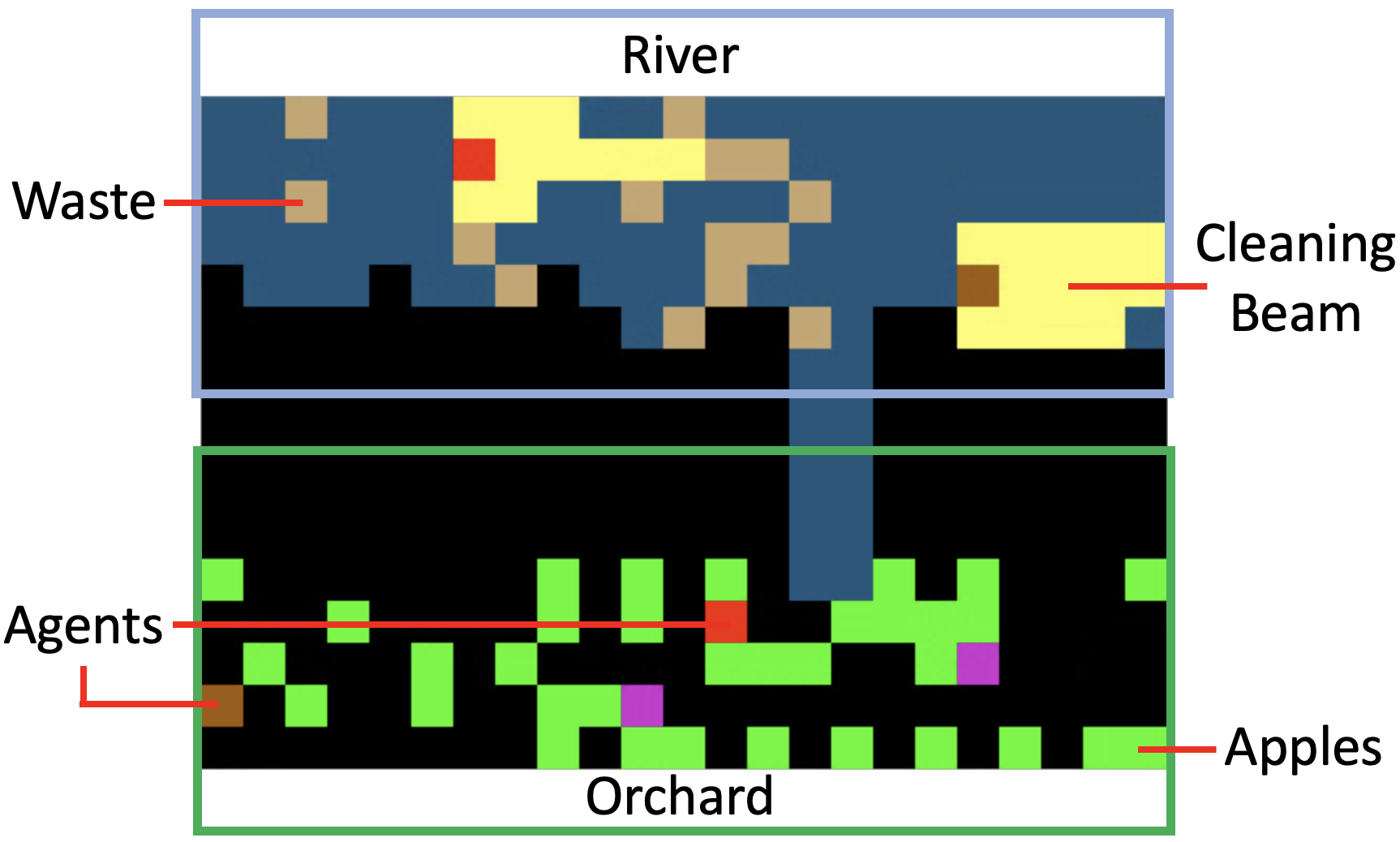}
    \caption{Screenshot of the Cleanup environment with three teams of two agents.}
    \label{fig:cleanup_env}
\end{figure}

\section{Equilibrium Analysis}
\label{sec:appendix_eq_anal}

When calculating the expected values of cooperation and defection with different credo, the fully self-focused and system-focused values are simply calculated using Table~\ref{tab:pd_table} of the main text.
Team-focused credo becomes more complex since it is a mixture of the mixed-motive and common interest game depending on the probability of being paired with a teammate $\inteam$.
We show the derivation for team-focused agents and continue with the final equilibrium with credo below.

\subsection{Team-Focused Agents}

Let $\sigma_{T_i}=(\sigma_{ji},1-\sigma_{ji})$ where $\sigma_{ji}$ is the probability for action $C$ represent when $i$ and $j$ have common interest (i.e., $j \in T_i$) and $\sigma_{T_j}=(\sigma_{jj},1-\sigma_{jj})$ when $i$ and $j$ do not have common interest (i.e., $j \in T_{j}$).
We now use common interest instead of strictly in the same team to scale to the self- and system-focus settings.
The expected utility of choosing to cooperate ($C$) or defect ($D$) for an agent with team-focused credo can be derived based on Table 1, $\inteam$, and the strategy of $j$ ($\sigma_{T_i}$ or $\sigma_{T_j}$).
First we show the derivation for a fully team-focused agent $i$'s expected utility for choosing $C$ subject to $j$'s strategy:


\begin{align}
   \mathbb{E}(C,\mathbf{\sigma}_T)_{\mathcal{T}} &= \inteam \left[ \sigma_{ji} (b-c) + (1-\sigma_{ji}) \frac{b-c}{2} \right] + \nonumber \\ &(1 - \inteam) \left[ \sigma_{jj} (b-c) + (1-\sigma_{jj}) -c \right]   \\
   &= \inteam \left[ \frac{2\sigma_{ji} (b-c)}{2} + \frac{b-c}{2} - \frac{\sigma_{ji} (b-c)}{2} \right] + \nonumber \\ &(1 - \inteam) \left[ \sigma_{jj} b - \sigma_{jj} c - c + \sigma_{jj} c \right] \\
   & = \inteam \left[ \frac{\sigma_{ji} b - \sigma_{ji} c}{2} + \frac{b-c}{2} \right] + (1 - \inteam) \left[ \sigma_{jj} b - c \right]\\
    & = \inteam \left[ \frac{(b - c) (\sigma_{ji} + 1)}{2} \right] + (1 - \inteam) \left[ \sigma_{jj} b - c \right] \\
    & =\frac{\inteam (b-c)(\sigma_{ji}+1)}{2}+ (1-\inteam)(\sigma_{jj} b-c).
\end{align}






Now we show the derivation for a team-focused agent $i$'s expected utility for choosing $D$ subject to $j$'s strategy:

\begin{align}
   \mathbb{E}(D,\mathbf{\sigma}_T)_{\mathcal{T}} &= \inteam \left[ \sigma_{ji} \frac{(b-c)}{2} \right] + (1 - \inteam) \left[ \sigma_{jj} b \right] \\
   &= \frac{\inteam \sigma_{ji}(b-c)}{2} + (1-\inteam)\sigma_{jj} b
\end{align}



The terms for playing defection with a counterpart who mutually defects ($1-\sigma_j$) is zero, and therefore omitted above.
Next, we show how the final equilibrium is derived using our parameters which define credo.

\subsection{Equilibrium with Credo}

Our credo vector defines how self-focused, team-focused, or system-focused an agent is while it learns in our environment.
We can calculate and derive when an agent has the incentive to cooperate in the Prisoner's Dilemma stage-game as:

\begin{equation}
\begin{multlined}
    \selfw \mathbb{E}(C,\mathbf{\sigma}_{T})_\mathcal{I} + \teamw \mathbb{E}(C,\mathbf{\sigma}_{T})_\mathcal{T} + \sysw \mathbb{E}(C,\mathbf{\sigma}_{T})_\mathcal{S} \geq \nonumber \\ 
    \selfw \mathbb{E}(D,\mathbf{\sigma}_{T})_\mathcal{I} + \teamw \mathbb{E}(D,\mathbf{\sigma}_{T})_\mathcal{T} + \sysw \mathbb{E}(D,\mathbf{\sigma}_{T})_\mathcal{S}.
    \label{eq:teams_constraint}
\end{multlined}
\end{equation}

Expanding each term with the derivations above and in the main text, we get:


\begin{align}
    &\selfw_i \left[ \sigma_{jj} (b - c) + (1 - \sigma_{jj}) (-c) \right] + \nonumber \\
    &\teamw_i \left[ \frac{\inteam (b-c)(\sigma_{ji}+1)}{2}+ (1-\inteam)(\sigma_{jj} b-c) \right] + \nonumber \\
    &\sysw_i \left[ \sigma_{ji} (b - c) + (1 - \sigma_{ji}) \left(\frac{b-c}{2} \right) \right] \geq
    \selfw_i \left[ \sigma_{jj} (b) \right] + \\
    &\teamw_i \left[ \frac{\inteam \sigma_{ji}(b-c)}{2} + (1-\inteam)\sigma_{jj} b \right] +
    \sysw_i \left[ \sigma_{ji} \left(\frac{b-c}{2} \right) \right]. \nonumber
\end{align}

We expand and simplify:
\begin{align}
    &\selfw_i \left[ \sigma_{jj} b - \sigma_{jj} c - c + \sigma_{jj} c \right] + 
    \teamw_i \left[ \frac{\inteam (b-c)(\sigma_{ji}+1)}{2} - c + \inteam c \right] + \nonumber \\
    &\sysw_i \left[ \sigma_{ji} (b - c) + \frac{b-c}{2} - \sigma_{ji} \left(\frac{b-c}{2} \right) \right] \geq 
    \selfw_i \left[ \sigma_{jj} (b) \right] + \\
    &\teamw_i \left[ \frac{\inteam \sigma_{ji} (b-c)}{2} \right] + 
    \sysw_i \left[ \sigma_{ji} \left(\frac{b-c}{2} \right) \right]. \nonumber
\end{align}
We can subtract everything on the right and be left with zero.

\begin{align}
    &\selfw_i \left[ -c \right] +
    \teamw_i \left[ \frac{\inteam (b-c)}{2} - c + \inteam c \right] + \nonumber \\
    &\sysw_i \left[ \sigma_{ji} (b - c) + \frac{b-c}{2} - \sigma_{ji} (b - c) \right] \geq 0,
\end{align}

\begin{equation}
    \selfw_i \left[ -c \right] +
    \teamw_i \left[ \frac{\inteam (b-c)}{2} - c + \inteam c \right] +
    \sysw_i \left[ \frac{b-c}{2} \right] \geq 0.
\end{equation}


The self- and system-focus terms are now fully simplified leaving the team-focus derivation remaining.
We move $\teamw_i \left[ 2c \right]$ to the other side of the inequality and simplify further.

\begin{equation}
    - \selfw_i c + 
    \teamw_i \left[ \inteam (b-c) + 2 \inteam c \right] +
    \sysw_i \left[ \frac{b-c}{2} \right] \geq 
    \teamw_i \left[ 2c \right]
\end{equation}

\begin{equation}
    - \selfw_i c + 
    \teamw_i \left[ \inteam - \frac{2c}{b+c} \right] +
    \sysw_i \left[ \frac{b-c}{2} \right] \geq 
    0
\end{equation}

\begin{equation}
    \teamw_i \left( \inteam - \frac{2c}{b+c} \right) +
    \sysw_i \left( \frac{b-c}{2} \right) \geq 
    \selfw_i c
    \label{eq:final_derivation}
\end{equation}

This last step represents the final derivation shown as Equation~\ref{eq:final_requrement} of the main text.
This equilibrium signifies under which conditions an agent has more incentive to cooperate than to defect.

\end{document}